\documentclass[journal]{IEEEtran}
\usepackage{graphicx}
\usepackage{amsmath,amssymb} 
\usepackage{color}
\usepackage{epsfig}
\usepackage{algpseudocode}
\usepackage{comment}
\usepackage[ruled,vlined]{algorithm2e}
\usepackage{subfig}

\usepackage{booktabs, multirow} 
\usepackage{tabu}
\usepackage{ntheorem}

\usepackage{rotating} 
\usepackage{soul}
\usepackage[table]{xcolor} 
\usepackage{hyperref}

\newcommand{\matr}[1]{\mathbf{#1}}

\newcommand{\floor}[1]{\left\lfloor #1 \right\rfloor}
\newcommand{\ceil}[1]{\left\lceil #1 \right\rceil}

\newcommand{\change}[1]{\textcolor{black}{#1}}

\begin{document}
%
\title{SemiCurv: Semi-Supervised Curvilinear Structure Segmentation}
%
%
%

\author{Xun Xu,
        Manh Cuong Nguyen,
        Yasin Yazici,
        Kangkang Lu,
        Hlaing Min,
        Chuan-Sheng Foo
\thanks{X. Xu, C. Nguyen, Y. Yazici, K. Lu, H. Min, and C.S. Foo are with the Institute for Infocomm Research (I2R), A-STAR, Singapore. e-mail: {xux}@i2r.a-star.edu.sg.}
}

%
%

\markboth{Journal of \LaTeX\ Class Files,~Vol.~14, No.~8, August~2015}%
{Shell \MakeLowercase{\textit{et al.}}: Bare Demo of IEEEtran.cls for IEEE Journals}
%



\maketitle

\begin{abstract}
   Recent work on curvilinear structure segmentation has mostly focused on backbone network design and loss engineering. The challenge of collecting labelled data, an expensive and labor intensive process, has been overlooked. While labelled data is expensive to obtain, unlabelled data is often readily available. 
   In this work, we propose SemiCurv, a semi-supervised learning (SSL) framework for curvilinear structure segmentation that is able to utilize such unlabelled data to reduce the labelling burden. Our framework addresses two key challenges in formulating curvilinear segmentation in a semi-supervised manner. First, to fully exploit the power of consistency based SSL, we introduce a geometric transformation as strong data augmentation and then align segmentation predictions via a differentiable inverse transformation to enable the computation of pixel-wise consistency. Second, the traditional mean square error (MSE) on unlabelled data is prone to collapsed predictions and this issue exacerbates with severe class imbalance (significantly more background pixels). We propose a N-pair consistency loss to avoid trivial predictions on unlabelled data. 
   We evaluate SemiCurv on six curvilinear segmentation datasets, and find that with no more than $5\%$ of the labelled data, it achieves close to $95\%$ of the performance relative to its fully supervised counterpart. The demo code is available at the project website~\footnote{\url{https://github.com/alex-xun-xu/SemiCurv}.}

\end{abstract}


%
\IEEEpeerreviewmaketitle

\section{Introduction}
Curvilinear structure segmentation aims to extract thin, curvilinear structures from images. It has wide applications, including road crack segmentation, road segmentation from satellite images and biomedical image segmentation (e.g. segmenting blood vessels and cells). 
Existing research into curvilinear structure segmentation has focused on designing better network architectures \cite{yang2019feature,wang2019context} or loss functions (e.g. topology loss \cite{hu2019topology,mosinska2018beyond}) to better account for the specific nature of curvilinear structures like connectivity and loop formation.

A major challenge in curvilinear structure segmentation that has received less attention is the need for a large amount of labelled data to obtain good performance. In many cases, acquiring such labelled data for segmentation tasks is non-trivial and sometimes requires expert labelling. Moreover, the fine detail present in curvilinear structures demands increased annotation effort in comparison to the coarse-grained polygon-based annotation that is sufficient for general (object-based) segmentation tasks. At the same time, we often have access to unlabelled data at very low cost. For example, road scanning vehicles are running everyday to collect road surface photographs, satellites are collecting aerial images on a daily basis, and medical images are abundant from daily diagnoses. This availability of unlablled data suggests the use of semi-supervised learning (SSL) approaches that are able to exploit the unlabelled data to reduce the amount of labelled data required. 

Much work in SSL has focused on semi-supervised classification \cite{van2020survey}, and  
far fewer works have studied the use of SSL for image segmentation, much less curvilinear structure segmentation.
Existing SSL methods for semantic segmentation adopt one of two approaches: 1) utilizing a Generative Adversarial Network (GAN) to enforce that predicted segmentation maps are indistinguishable from real segmentation maps
\cite{hung2019adversarial,mondal2019revisiting} 
or 2) enforcing consistency of predicted segmentation maps on unlabelled data between two augmented samples \cite{ouali2020semi,french2019semi,olsson2021classmix}. The first strategy still requires a considerable amount of labelled data to train the GAN. 
In contrast, consistency-based approaches are able to work with much less labelled data and are more promising in the low-label regime.  Therefore, we build upon the consistency-based approach towards semi-supervised segmentation in this work. 


Even though previous work has developed methods for semi-supervised segmentation, we argue that direct application of these methods to the curvilinear segmentation task is sub-optimal for the following two reasons. {First, the types of data augmentation used by state-of-the-art SSL methods for generic segmentation tasks has thus far been limited to pixel-wise perturbations, including a mix-up of multiple images~\cite{french2019semi,olsson2021classmix}. Pixel-wise perturbations do not create more diverse geometric views of image data (e.g. vertical cracks will always remain vertical under pixel-wise perturbations), and thus may not fully exploit the power of consistency based SSL. Moreover, mixing up images could shift the data away from the real data manifold (i.e. the mixed-up image may not look real), potentially harming SSL as demonstrated in our experiments. Secondly, curvilinear structures typically occupy a small fraction (often under 10\% and as little as 1\% in the datasets we considered) of the image compared to the background, leading to a severely class-imbalanced segmentation target; this is unlike the more general semantic segmentation datasets~\cite{everingham2015pascal} considered by these existing methods. As we will show, the class imbalance and sparsity leads to a ``collapsing'' issue with the consistency loss used by SSL methods, resulting in models predicting constant outputs.

We next describe the key improvements we make to consistency-based SSL to address these challenges associated specifically with curvilinear segmentation tasks: 

\emph{1) Differentiable affine transformation.} First, we propose affine transformations as data augmentation for curvilinear images.
This augmentation is effective because of two reasons. First, images with curvilinear structures are often captured from planar surfaces; this is characteristic of images of road surfaces, satellite images, and images of cells taken under a microscope, to give a few examples. Hence, synthesizing novel viewpoints as augmentation by affine transformation is valid~\cite{andrew2001multiple}. Second, segmentation of curvilinear structures is expected to be affine equivariant: for instance, the notion of a crack on a road surface should remain the same regardless of rotation, translation or resizing of the image, unlike in  semantic segmentation tasks where the pose of an semantic object is often fixed or subject to small variation. 

One challenge with using affine transformations as data augmentation is that 
despite being more diverse, they prohibit computing consistency loss at the pixel level between two randomly augmented images because pixel-wise correspondence is not preserved during affine transformation. As illustrated in Fig.~\ref{fig:Interpolation}, supposing $\matr{\hat{I}}$ and $\matr{\bar{I}}$ are randomly augmented from $\matr{I}$, there is no explicit pixel-wise correspondence between $\matr{\hat{I}}$ and $\matr{\bar{I}}$ with which we can enforce the predictions to be consistent. 
To address this alignment issue, we propose to employ an inverse affine transformation following the networks'  predictions as illustrated in Fig.~\ref{fig:Framework}. With both forward and inverse transformations, the pixel-to-pixel correspondence is restored, enabling the computation of the pixel-wise consistency loss between student and teacher models. We note that incorporating the inverse transformation into the networks requires it to be differentiable for gradients to backpropagate; as affine transformation is differentiable, the inverse transformation allows end-to-end learning as part of the networks.

\begin{figure}
    \centering
    \includegraphics[width=1.0\linewidth]{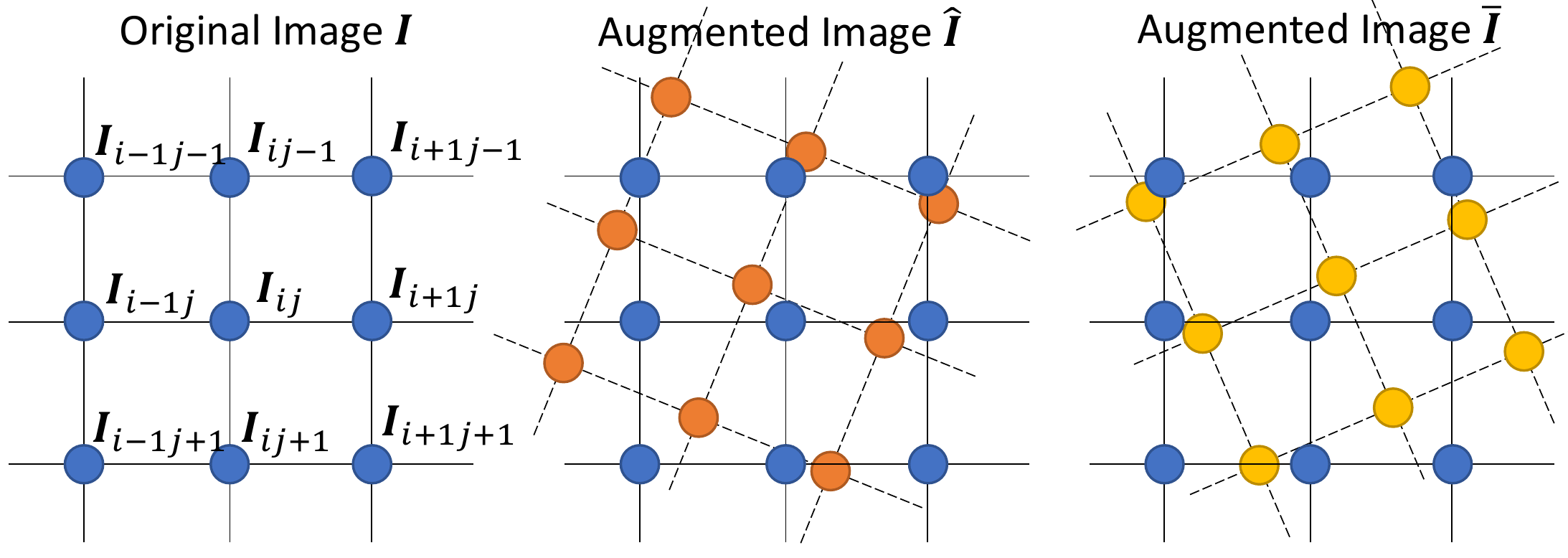}
    \caption{Illustration of aligning augmented images. Two random augmentations are applied to original image $\matr{I}$, resulting in $\matr{\hat{I}}$ and $\matr{\tilde{I}}$. As pixel-to-pixel correspondence is no longer available between $\matr{\hat{I}}$ and $\matr{\tilde{I}}$, we apply additional inverse transformation to align two augmented images.}
    \label{fig:Interpolation}
\end{figure}



} 

\emph{2) N-pair consistency.} As noted above, we observe that the widely adopted consistency loss, mean square error (MSE), is prone to a ``collapsing'' issue. Trivial solutions to minimize the MSE on unlabelled data exist where both student and teacher networks predict constant outputs. The introduction of a non-learnable teacher model may partially address this issue, but cannot guarantee in principle that a trivial solution is not learned. Moreover, we notice that there are often far fewer positive (foreground) pixels than negative ones (background) in curvilinear structure datasets, suggesting the prior distribution is highly biased towards background. This imbalance further encourages the trivial solution where the learned model predicts all background on the unlabelled data. To overcome this challenge, we propose to use an N-pair loss \cite{sohn2016improved} to avoid trivial predictions on unlabelled data, inspired by the recent success of contrastive learning \cite{chen2020simple,he2020momentum}.




\emph{3) Spatial coordinate encoding.} Finally, there is clear spatial connectivity and correlation in curvilinear structure patterns, in that positive pixels are spatially adjacent. Although topology was studied to improve connectivity \cite{mosinska2018beyond,hu2019topology}, they require fully labelled ground-truth to provide topological supervision. To exploit the spatial correlation prior, we propose to add spatial coordinate encoding as features to all layers of the backbone segmentation network. To avoid the potential overfitting to absolute spatial coordinates \cite{liu2018intriguing}, we encode spatial adjacency using a sinusoid function, which preserves spatial adjacency information while reducing risk of overfitting. 

In summary, our work makes the following contributions:
\begin{itemize}
    \item We introduce a differentiable affine transformation to fully exploit the strength of consistency based semi-supervised learning for curvilinear image segmentation. 
    \item We provide insight into a ``collapsing'' issue with MSE consistency loss defined on unlabelled data and propose to use N-pair loss to mitigate this issue.
    \item To further capture the spatial correlation, we incorporate sinusoid spatial encoding as additional features.
    \item To the best of our knowledge, this is the first attempt to investigate into curvilinear structure segmentation in a semi-supervised fashion. We extensively benchmarked state-of-the-art semi-supervised methods on six curvilinear segmentation datasets.
\end{itemize}

\section{Related Work}
\subsection{Curvilinear Structure Segmentation}


We briefly review previous work on curvilinear structure segmentation in the context of road segmentation from satellite images, medical image segmentation and crack detection in pavements.
Early work in road segmentation \cite{mnih2010learning} used fully connected networks, but were soon followed by CNNs \cite{MnihThesis}. Recent work has built on the CNN approach, focusing on designing deeper and wider networks and considering information at multiple scales \cite{henry2018road, liu2018roadnet}. In the medical imaging community, efforts have focused on further improving the popular UNet backbone ~\cite{ronneberger2015u, isensee2019nnu}. These fully supervised methods have shown to be competitive in delineating linear structures like membrane segmentation \cite{arganda2015crowdsourcing}. Progress in the area of crack segmentation proceeded in a similar trajectory, where recent progress is due to improved backbone network design \cite{zou2018deepcrack,yang2019feature}. Orthogonal to the above, in this work, we address the semi-supervised learning setting to reduce the amount of labelled data needed to train high performance models by exploiting unlabelled data.

To model spatial correlation and connectivity in curvilinear structures, a few works have incorporated topological constraints when training the segmentation model. One study utilized intermediate features from a VGG network to capture topological structure \cite{mosinska2018beyond}, while another work used concepts from persistent homology to provide more principled topological features; this idea 
was implemented by defining a loss between ground-truth and predicted persistent diagrams \cite{hu2019topology}, which assumes access to ground-truth labels. 
In this work, we adopt a simpler approach by using positional encodings to capture spatial correlation. 

\subsection{Semi-Supervised Learning}

We briefly review the two genres of semi-supervised learning (SSL) methods, namely the consistency-based SSL and adversarial training based SSL; for a more complete review, please refer to \cite{van2020survey}. Consistency-based SSL methods \cite{laine2016temporal,tarvainen2017mean,miyato2018virtual} utilize the unlabelled data to constrain learning by enforcing the model's predictions on the unlabelled data to be consistent with a specified target. 
Consistency regularization was firstly introduced in the context of deep learning by \cite{rasmus2015semi}. 
To better exploit the power of ensemble learning \cite{laine2016temporal} used temporal ensembling of models' predictions as a target for consistency. \cite{tarvainen2017mean} further proposed to ensemble the model parameters over time to obtain more stable consistency targets. 
Different from the previously mentioned approaches, the VAT method \cite{miyato2018virtual} proposed to enforce consistency of a model's predictions with that on an adversarially perturbed input \cite{goodfellow2014explaining}.  
Orthogonal to the above, the MixMatch method \cite{berthelot2019mixmatch} proposed to interpolate between labelled and unlabelled samples to diversify the training set. 
Due to the high accuracy of temporal ensembling and low memory footprint, we adopted Mean Teacher \cite{tarvainen2017mean} as the backbone framework.
\change{Another line of works adapt generative adversarial training to semi-supervised learning. Specifically, in the seminal work~\cite{odena2016semi}, a discriminator was introduced to distinguish generated samples from real ones and at the same time classify real samples into respective categories.}

\subsection{Semi-Supervised Segmentation}

\change{
Semi-Supervised learning has been applied to semantic image segmentation to alleviate the expensive image annotation task. 
Following the adversarial training approach~\cite{odena2016semi}, \cite{souly2017semi} first employed a discriminator to distinguish generated fake images from real ones to improve semi-supervised learning. A conditional generative adversarial network (GAN) was further employed by ensuring that a discriminator cannot distinguish predicted segmentation masks on unlabelled images from ground-truth segmentation masks (on the labelled images) \cite{hung2019adversarial}; \cite{mondal2019revisiting} further imposes cycle consistency (building on CycleGAN) to constrain the learning on unlabelled data. However, under low labeled data training a GAN is prone to instability and may not generalize well to unlabeled data.
Following the consistency-based SSL approaches, \cite{french2019semi} proposed to enforce consistency between the segmentation predictions of images subject to a MixUp augmentation, i.e. random crops of the two images are superimposed and consistency is enforced on the segmentation predictions of respective patches. 
In the context of medical image segmentation, a framework similar to Mean Teacher has been applied to skin lesion and CT-scan segmentation \cite{li2020transformation}.
All these existing methods only consider data augmentations that are either pixel-wise perturbations \cite{french2019semi} or rotations that are multiples of $90^\circ$ \cite{li2020transformation} with pairwise consistency; such augmentations are not strong enough to fully exploit the power of consistency regularization. These works also do not address the issue where the MSE loss used to enforce consistency is prone to collapse.}

\begin{figure}[!htb]
    \centering
    \includegraphics[width=0.99\linewidth]{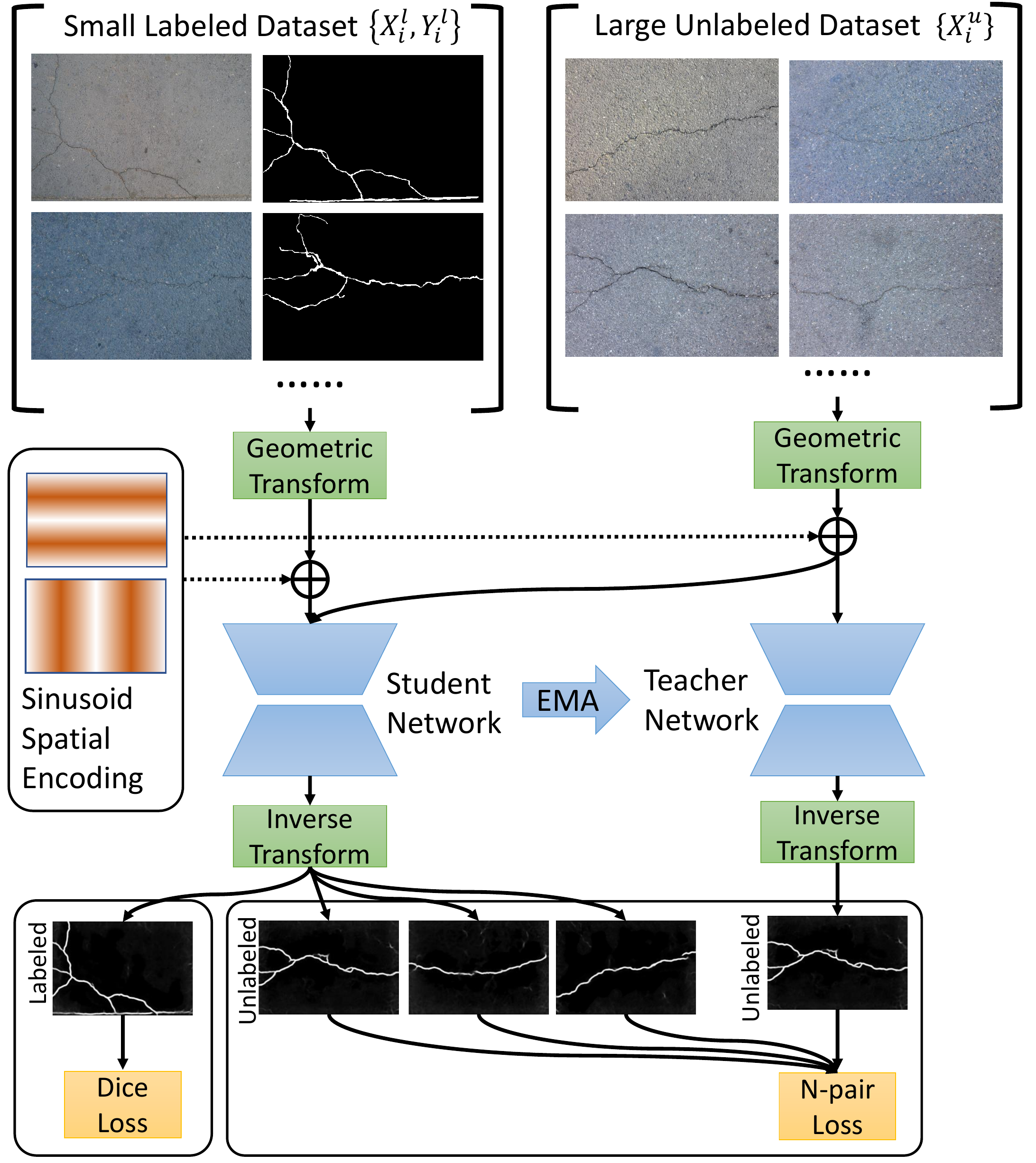}
    \caption{\change{Overview of SemiCurv, a semi-supervised learning framework for curvilinear structure segmentation. We adopt a mean-teacher like architecture for semi-supervised learning. i) We first propose a geometric transformation and the inverse transformation to allow stronger data augmentation. ii) To account for the severe class-imbalance issue, we employ Dice loss and N-Pair loss for labelled and unlabelled data respectively. iii) Spatial encoding is further incorporated to capture spatial correlations.} 
    }
    \label{fig:Framework}
\end{figure}

\section{Methodology}

\subsection{Overview of SemiCurv}
We first provide an overview of our SemiCurv approach, then describe the individual components in greater detail. Our SemiCurv approach, illustrated in Fig.~\ref{fig:Framework}, is built upon the consistency-based SSL framework (Sect.~\ref{sect:cons-ssl}). Both labelled and unlabelled images are first augmented by a differentiable geometric transformation (Sect.~\ref{sect:GeoTform}) and then fed into student and teacher networks. The teacher network is a parameter-wise temporal ensemble (exponential moving average) of the student network and generates a target (pseudo-label) to encourage consistency on the unlabelled data. Sinusoid positional encoding is appended to each convolution layer in the encoder to explicitly capture spatial correlations   (Sect.~\ref{sect:PositionEnc}). The predicted segmentation maps (posterior probabilities) from both networks are transformed back to the original pose via the differentiable inverse transform, and the dice loss (Sect.~\ref{sect:Dice}) and N-pair loss (Sect.~\ref{sect:Consistency}) are used on the labelled and unlabelled data respectively for training; in particular, we show how the N-pair loss resolves a ``collapsing'' issue associated with the MSE loss commonly used for consistency on the unlabelled data.

\subsection{Consistency-based Semi-supervised Learning}
\label{sect:cons-ssl}

We denote a training set of $N_l$ labelled images and their ground-truth segmentation maps as $\mathcal{D}_l=\{(\matr{X}^l_i,\matr{Y}^l_i)\}_{i=1}^{N_l}$, and denote the set of $N_u$ unlabelled images as $\mathcal{D}_u=\{\matr{X}^u_j\}_{j=1}^{N_u}$. 
The loss function optimized by semi-supervised learning methods can be generally   written as Eq.~(\ref{eq:ConsistSSL}), where $\mathit{l}_l$, $\mathit{l}_u$, and $\gamma$ are supervised loss on labelled images, unsupervised loss on unlabelled images and balancing hyperparameter, respectively. 

\begin{equation}\label{eq:ConsistSSL}
    L(\mathcal{D}_l, \mathcal{D}_u) = \frac{1}{N_l}\sum_{i=1}^{N_l}\mathit{l}_l(\matr{X}^l_i,\matr{Y}^l_i)+\gamma \frac{1}{N_u}\sum_{j=1}^{N_u} l_u(\matr{X}^u_j)
\end{equation}
{We note that the loss functions of multiple SSL methods \cite{tarvainen2017mean,laine2016temporal,miyato2018virtual} can be written in this form. Most existing works adopt the cross-entropy loss or its variants as the supervised loss $l_l$ for classification tasks. In this work, we demonstrate in Sect.~\ref{sect:Dice} that for segmentation with severe class imbalance, Dice loss is more appropriate. For the unsupervised loss, the $\Pi$ model \cite{laine2016temporal}, mean teacher model (MT) \cite{tarvainen2017mean}, VAT \cite{miyato2018virtual}, MixMatch \cite{berthelot2019mixmatch} and variants, all apply mean square error (MSE) consistency between the prediction posteriors of an unlabelled image under two different augmentations $t_1(\cdot)$ and $t_2(\cdot)$, and two different segmentation networks $f_1(\cdot)$ and $f_2(\cdot)$ as in Eq.~\ref{eq:MSE}. We show in Sect.~\ref{sect:Consistency} that under severe class imbalance the N-pair loss is more effective than MSE in avoiding trivial solutions.} 


\begin{equation}\label{eq:MSE}
    l_u(\matr{X}) = ||f_1(t_1(\matr{X}))-f_2(t_2(\matr{X}))||_F^2
\end{equation}


SemiCurv is based on the mean teacher (MT) method \cite{tarvainen2017mean} for semi-supervised classification, which has also been previously applied to the segmentation task. Briefly, the MT method maintains two networks: one fully trainable network called the student network, and another non-trainable network called the teacher network. The teacher network provides pseudo-labels to supervised the student network to learn. To improve stability of training, the parameters of teacher are the exponential moving average (EMA) of the student network's parameters. In the rest of the paper, we denote the student network as $f(\cdot)$ and teacher network as $\hat{f}(\cdot)$.


\begin{figure*}[!htb]
    \centering
    \subfloat[]{
    \includegraphics[width=0.77\linewidth]{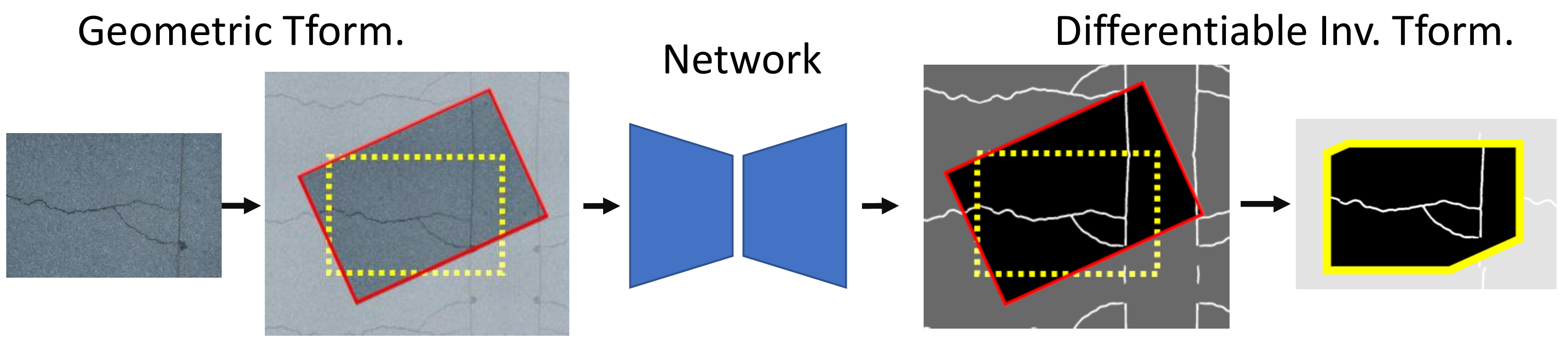}
    }
    \subfloat[]{\includegraphics[width=0.22\linewidth]{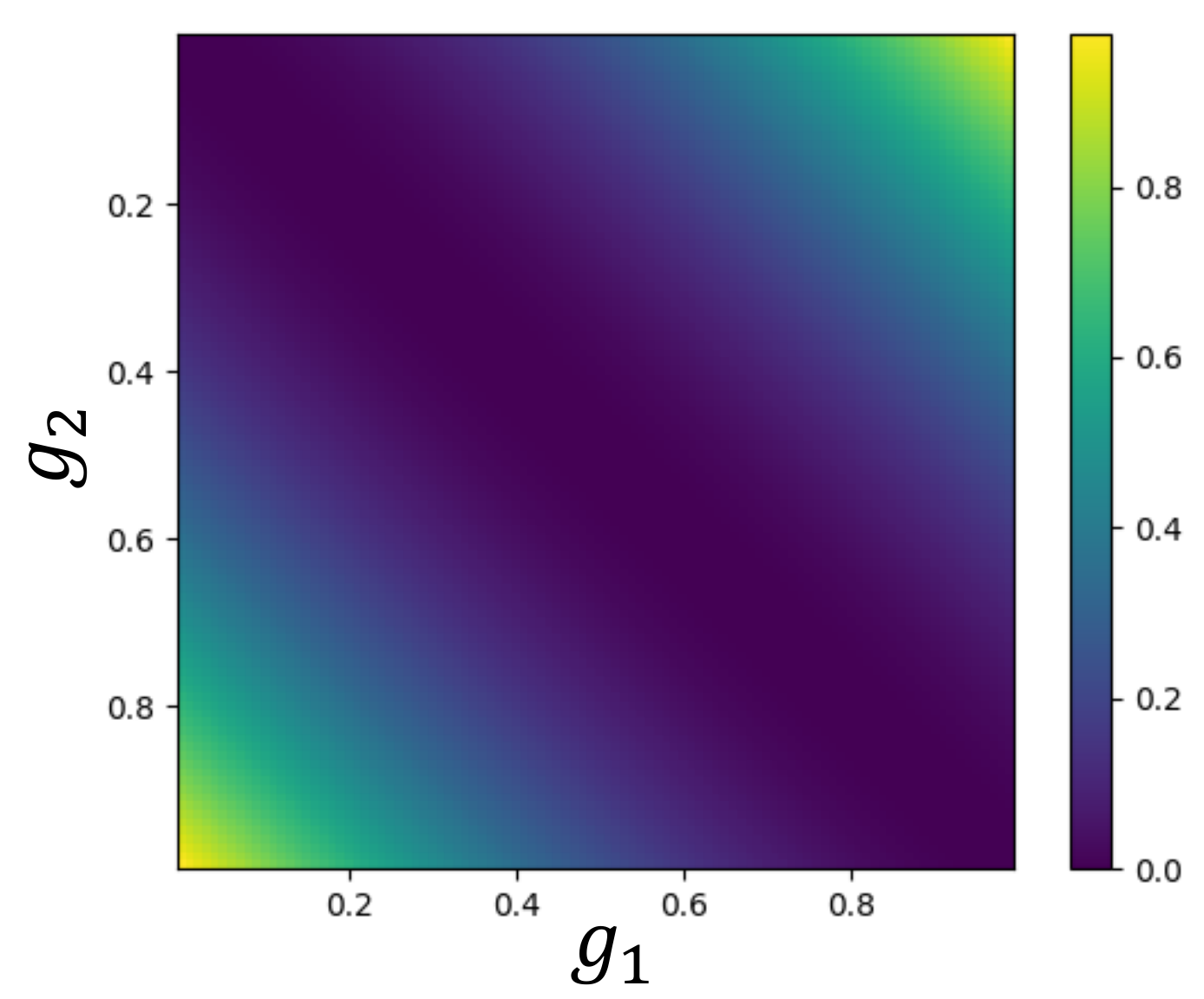}}
        \vspace{-0.3cm}
    \caption{(a) Illustration of differentiable transformation. Red solid box and yellow dashed box indicate transformed and original views. The final yellow solid box indicates the prediction after differentiable inverse transformation where consistency loss is computed. (b) Loss surface for MSE consistency loss. When both student and teacher produce exactly the same result (may not be correct), as indicated by the diagonal, the MSE loss is always zero.}
    \vspace{-0.5cm}
    \label{fig:AugAndNSE}
\end{figure*}

\subsection{Differentiable Geometric Transformation for Augmentation}\label{sect:GeoTform}

Strong data augmentation is key to the success of consistency-based semi-supervised learning \cite{xie2020unsupervised}. As discussed, existing semi-supervised segmentation methods often consider a limited set of augmentations. {To further increase the variation of poses to improve feature learning we introduce affine transformations for stronger data augmentation. We denote the transformation applied to input image $\matr{X}$ as $t(\matr{X})$. To enable computing pixel-wise consistency loss, we further apply the inverse transformation to the output of both student and teacher networks as $t^{-1}(f(t(\matr{X})))$. With the inverse transformation, predictions on two arbitrarily augmented images are directly comparable at the
pixel level. 
} 


Concretely, an affine transformation involves an arbitrary combination of scaling, rotation, shearing and translation, and can be formulated as a matrix multiply with the following transformation matrix,
\begin{equation}
    \matr{H}=\begin{bmatrix}
a_{11} & a_{12} & d_x\\
a_{21} & a_{22} & d_y \\
0 & 0 & 1
\end{bmatrix}
\end{equation}

For every pixel at $(\hat{w},\hat{h})$ in the image after affine transformation, we find its coordinate in the original image by $[w,h,1]^\top=\matr{H}^{-1}[\hat{w},\hat{h},1]^\top$. 
We  employ bilinear interpolation to produce the pixel intensity at every pixel after transformation.  Concretely, we can find the 4 neighbouring pixels before the transform as $\{x_{\floor{w},\floor{h}},x_{\ceil{w},\floor{h}},x_{\floor{w},\ceil{h}},x_{\ceil{w},\ceil{h}}\}$. 
The pixel intensity value after transformation $\hat{x}_{\hat{w},\hat{h}}$ can be determined by bilinear interpolation as,
\begin{equation}
    \hat{x}_{\hat{w},\hat{h}} = [\ceil{w}-w,w-\floor{w}]
    \begin{bmatrix}
    x_{\floor{w},\floor{h}} & x_{\ceil{w},\floor{h}}\\
    x_{\floor{w},\ceil{h}} & x_{\ceil{w},\ceil{h}}
    \end{bmatrix}
    \begin{bmatrix}
    \ceil{h}-h\\
    h-\floor{h}
    \end{bmatrix}
\end{equation}

Because of the bilinear transformation $\hat{x}_{\hat{w},\hat{h}}$ is obviously differentiable w.r.t. $\{x_{w,h}|w\in\{0,W-1\},h\in\{0,H-1\}$ where $H$ and $W$ are the height and width of image respectively, i.e. $t(\matr{X})$ is differentiable w.r.t. $\matr{X}$. For the inverse transformation, we simply apply $\matr{H}^{-1}$ for affine transformation 
and thus $t^{-1}(f(t(\matr{X})))$ is still differentiable w.r.t. $\matr{X}$. 

We also notice that to generate diverse and realistic augmented image samples, we extrapolate images by mirroring the image over the edges. To give an example, the result of a 1D sequence ``$abcd$'' after mirror flipping is ``$dcb|abcd|cba$''. This is achieved by clipping the pixel location before transformation with $x=(x//W\%2)*(W-x\%W)+(1-x//W\%2)*(x\%W)$ where $//$ and $\%$ are floor and modulo divisions respectively. An illustration of augmentation for image segmentation is given in Fig.~\ref{fig:AugAndNSE}~(a). With the differentiable transformation, we are able to randomly generate two augmented images $t_1(\matr{X})$ and $t_2(\matr{X})$ that are respectively fed into the student and teacher networks $f$ and $\hat{f}$. The two prediction posteriors can be then aligned via the differentiable inverse transformation $t_1^{-1}(f(t_1(\matr{X})))$ and $t_2^{-1}(\hat{f}(t_2(\matr{X})))$. For the rest of the paper, we denote prediction posteriors after alignment as $g_{i1}=t_1^{-1}(f(t_1(\matr{X}_i)))$ (student) and $\hat{g}_{j2}=t_2^{-1}(\hat{f}(t_2(\matr{X}_j)))$ (teacher).




\subsection{Avoiding Collapsed Predictions on Unlabelled Data}
\label{sect:Consistency}



Consistency-based SSL methods commonly use mean square error (MSE) for the consistency loss on unlabelled data. In this section, we first point out that MSE loss allows collapse of model predictions on unlabelled data to the majority class as this is a trivial solution of the MSE consistency loss. We then introduce the N-pair loss as a way to mitigate this issue thus enabling the unlabelled data to better regularize the model.

Without loss of generality, suppose we have two scalar predictions (pixel-wise predictions) $g_1$ and $g_2$. We visualize the loss surface for MSE pairwise consistency in Fig.~\ref{fig:AugAndNSE}~(b), and observe that the MSE loss is flat along the diagonal because MSE loss is minimized when the predictions from the two networks match. When the class distribution is highly imbalanced as in the case of curvilinear segmentation, where there are far more background pixels than foreground pixels, the two networks can achieve zero MSE consistency loss just by assigning every single pixel to the majority class. In this case every pixel will be predicted as background. We term this all majority class prediction as a collapsed prediction. An instance of this behaviour is shown in Fig.~\ref{fig:CollapseSinusoidPeriod}~(a) where training IoU gets closer to $1$, yet validation IoU collapses to $0$. Avoiding this often requires very careful selection of the EMA hyperparameter and consistency weights, which requires expensive tuning runs. 

\begin{figure}[htb]
    \centering
    \includegraphics[width=1\linewidth]{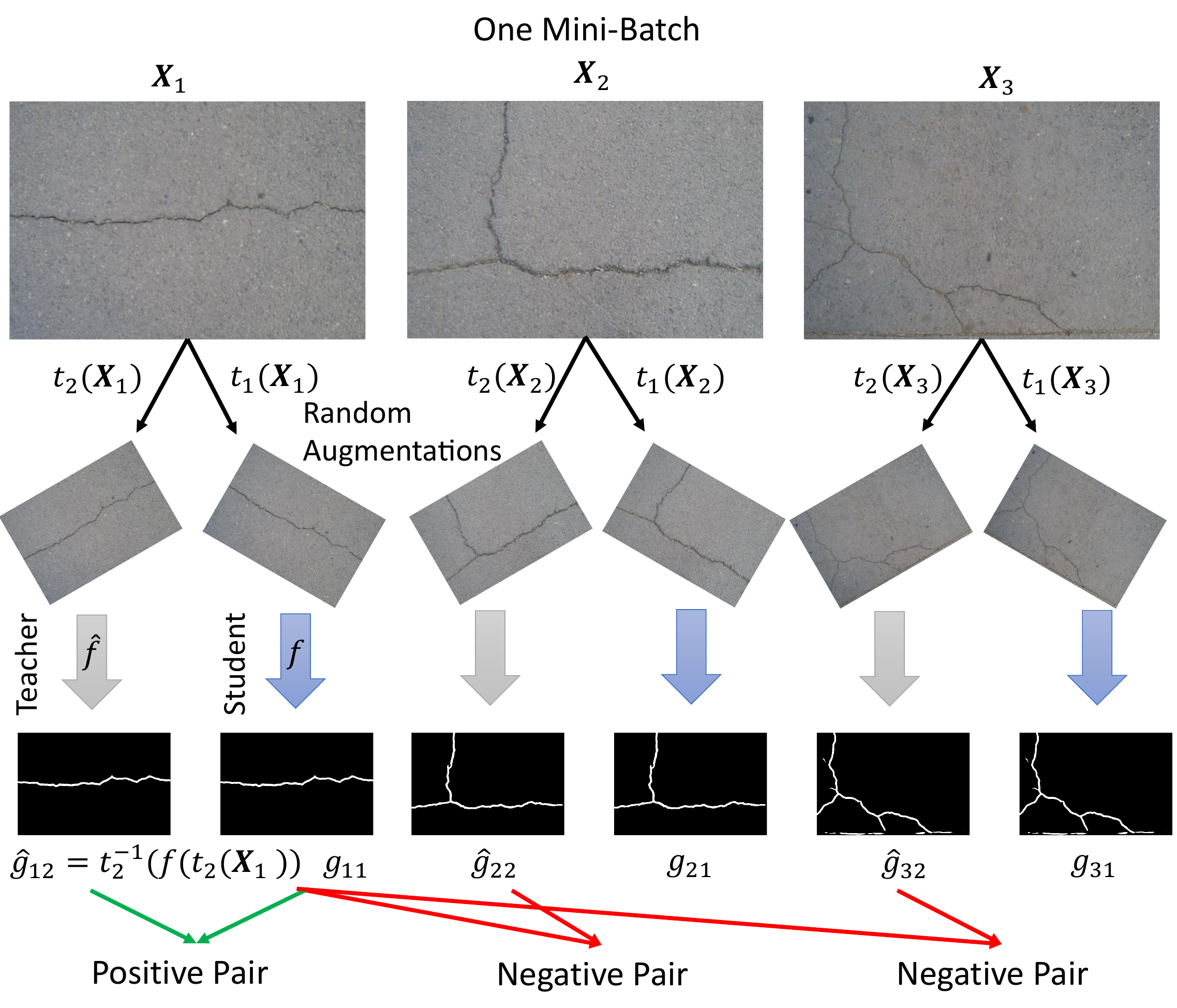}
    \caption{\change{Illustration of N-pair loss for unlabelled images. Positive pairs are constructed between the predictions of teacher and student networks on the same input image. Negative pairs are constructed between the predictions of teacher and student networks on different input images.}}
    \vspace{-0.5cm}
    \label{fig:ContrastiveLoss}
\end{figure}

\change{In this work, inspired by the recent progress in contrastive learning \cite{chen2020simple}, we propose to use an N-pair loss on unlabelled data~\cite{sohn2016improved} to avoid collapsed predictions. The N-pair loss simultaneously exploits all unlabelled samples in a training mini-batch to construct one positive pair to encourage similarity and multiple negative pairs to encourage diversity in predictions. By enforcing diversity with the negative pairs, the N-pair loss effectively penalizes models that give collapsed predictions on all unlabelled data. We illustrate the construction of the N-pair loss for curvilinear structure segmentation in Fig.~\ref{fig:ContrastiveLoss}: in a mini-batch of $N_B$ unlabelled images 
the positive pair is chosen as the predictions from student and teacher on an anchor image $\matr{X}_i$, $\{(g_{i1},\hat{g}_{i2})|i=1\cdots N_B\}$, while negative pairs are chosen as the predictions between the student's predictions on the same anchor image and teacher's predictions on other images in the mini-batch, $\{(g_{i1},\hat{g}_{j2})|i\neq j; i,j=1\cdots N_B\}$. Formally, the N-pair loss is given by,}

\begin{equation}\label{eq:N-pairLoss}
    l_{\textrm{N-pair}}=-\frac{1}{N_B}\sum_{i=1}^{N_B}\log\frac{\exp(sim(g_{i1},\hat{g}_{i2})/\tau)}{\sum_{j=1}^{N_B} \exp(sim(g_{i1},\hat{g}_{j2})/\tau)}
\end{equation}

\noindent where $N_B$, $\tau$ and $sim$ are the mini-batch size, temperature parameter and similarity metric, respectively. We use cosine similarity defined in Eq.~(\ref{eq:CosineSim}) for the similarity metric $sim$, where $\mathbf{vec}$ vectorizes a matrix into a vector and $\delta$ is a small value to avoid division by zero. 
\begin{equation}\label{eq:CosineSim}
\resizebox{0.75\linewidth}{!}{
$
    sim({g}_1,{g}_2)=\frac{\mathbf{vec}({g}_1 +\delta)^\top \mathbf{vec}({g}_2+\delta) }{||\mathbf{vec}({g}_1+\delta)||_2\cdot||\mathbf{vec}({g}_2+\delta)||_2}
$
}
\end{equation}

{We now formally show how N-pair loss prevents collapsed predictions that can occur when MSE is used. Specifically, we show that the N-pair loss will be lower for correct predictions than collapsed predictions, unlike MSE loss. First consider the case when predictions all collapse to a single class, often the background resulting in all zero predictions, i.e. $g_1=\matr{0}$ and $\hat{g}_2=\matr{0}$. Then, the cosine similarity for all positive and negative pairs are 1; the same result holds when predictions are all foreground, i.e.  $g_1=\matr{1}$ and $\hat{g}_2=\matr{1}$. Then, the N-pair loss for this collapsed prediction is
\begin{equation}
 \tilde{l}_\textrm{N-pair}=-\frac{1}{N_B}*N_B\log 1/N_B=\log N_B.   
\end{equation}
We note that in this case, the MSE loss is $\tilde{l}_{mse}=0$. When the predictions are all correct, the positive pair similarity is $1$, and for negative pairs this is approximately $\epsilon<<1$, because the segmentation masks for arbitrary two images are unlikely to significantly match, and the N-pair loss becomes
\begin{equation}
\begin{split}
l_\textrm{N-pair}&=-\frac{1}{N_B}*N_B(\log \frac{\exp1/\tau}{\exp1/\tau+(N_B-1)*\exp\epsilon/\tau})\\
&=\log (1+(N_B-1)*\exp\frac{\epsilon-1}{\tau})
\end{split}
\end{equation}
while the MSE loss is still $l_{mse}=0$. Since $\exp\frac{\epsilon-1}{\tau}<1$, it is easy to verify that $l_\textrm{N-pair}<\tilde{l}_\textrm{N-pair}$ when $N_B$ is larger than 1 and the ratio $l_\textrm{N-pair}/\tilde{l}_\textrm{N-pair}$ is getting smaller with increased batchsize $N_B$. This suggests the gap between N-pair loss under collapsed prediction and correct prediction is more significant with larger batchsize. We therefore conclude that the N-pair loss can easily distinguish good predictions from collapsed ones as opposed to MSE loss, for which the loss is $0$ in both cases. }





\subsection{Modelling Spatial Correlation with Positional Encoding}\label{sect:PositionEnc}



\begin{figure}[htb]
    \centering
    \includegraphics[width=0.9\linewidth]{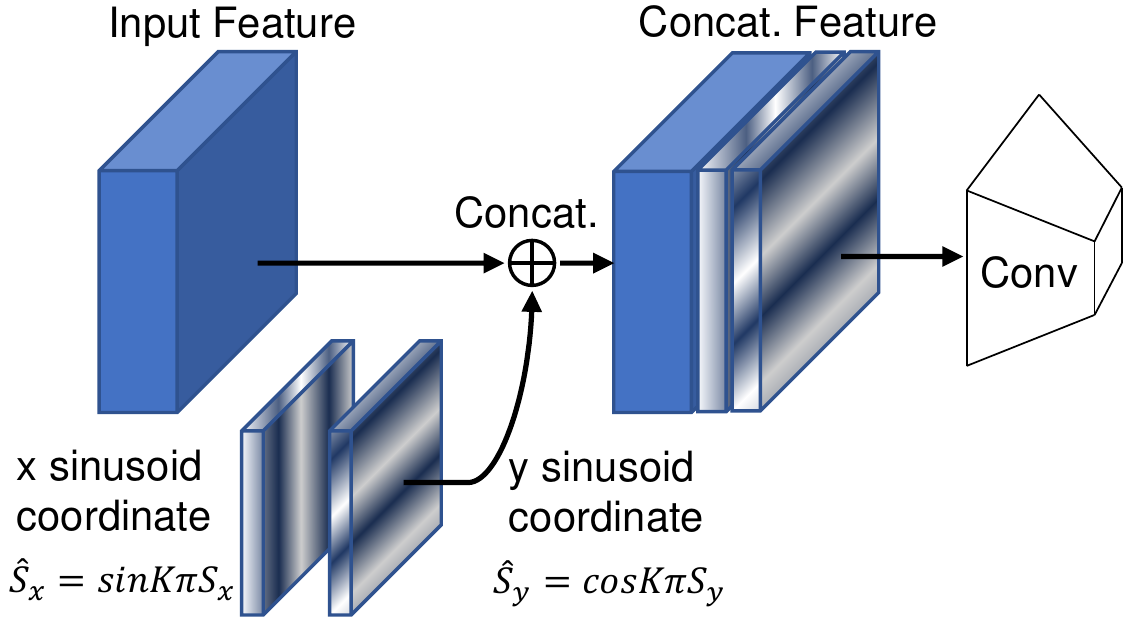}
    \caption{Sinusoid coordinate encoding for capturing spatial correlation. Encodings are concatenated to feature maps in channel-wise fashion.}
    \label{fig:SinusoidCoord}
\end{figure}

The nature of curvilinear structure segmentation requires the network to encode spatial correlation. For example, cracks, roads, and blood vessels are spatially continuous thus pixels adjacent to other positive (foreground) pixels are likely to be positive as well. Though it is difficult to handcraft features that capture this correlation, encoding spatial locations has been shown to be effective \cite{liu2018intriguing}. One straightforward way to encode spatial locations is by appending linear coordinates, normalized to between 0 and 1, to the intermediate feature layers as additional feature channels. However, learning directly on this absolute positional encoding risks overfitting to specific locations, especially when training with very few labelled samples.
Inspired by the positional encoding of \cite{vaswani2017attention}, we apply a sinusoid coordinate encoding as in Eq.~(\ref{eq:SinEnc}) where $S_{x}, S_{y}$ are the linear positional encoding normalized to between 0 and 1 and $K$ is a period parameter. Such periodic positional encoding allows the network to be aware of relative location while reducing the risk of overfitting to absolute locations. The positional encodings are channel-wise concatenated to each intermediate activation map of the encoder network as shown in Fig.~\ref{fig:SinusoidCoord}.

\begin{equation}\label{eq:SinEnc}
    \hat{S}_x=\sin K\pi S_x,\quad \hat{S}_y=\cos K\pi S_y
\end{equation}

\subsection{Dice Loss for Class Imbalance}
\label{sect:Dice}
It is well-known that class imbalance can affect the performance of machine learning models \cite{4667275,johnson2019survey,Guo2017LearningFC}. The cross-entropy (CE) loss commonly used for supervised segmentation tasks averages the pixel-wise CE loss over all pixels in the image; CE loss biases learning towards the majority class when there is an imbalanced class distribution. In curvilinear structure segmentation tasks, this imbalance is particularly severe due to a low positive class ratio (see Table~\ref{tab:PosRatio}). As a result, models trained with CE may have high per-pixel accuracy but low intersection over union (IoU), which is the most common evaluation metric for segmentation tasks. \change{In order to mitigate this issue, 
Dice loss~\cite{sudre2017generalised} (Eq.~\ref{eq:DiceLoss}) was proposed as an alternative for segmentation; it focuses on the intersection of predictions with ground truth only for the positive class.}

\begin{equation}\label{eq:DiceLoss}
    l_{dice}=1-\frac{2\mathbf{vec}(g_1+\delta)^\top\mathbf{vec}(\matr{Y}+\delta)}{||\mathbf{vec}(g_1+\delta)||_1+||\mathbf{vec}(\matr{Y}+\delta)||_1}
\end{equation}

\subsection{Implementation Details}
The final loss function is the weighted combination of supervised loss and unlabelled loss: $l_{total}=l_{dice}+\gamma l_\textrm{N-pair}$. As commonly done in consistency-based SSL, we do not apply the unlabelled loss $l_\textrm{N-pair}$ at the beginning of training but instead gradually increase the strength of $\gamma$ following a sigmoid function $\gamma=\exp(-10(\frac{min(T,T_{rp})}{T_{rp}}-1)^2)$ where $T$ and $T_{rp}$ are epoch number and hyperparameter, respectively. We set the mean teacher EMA hyperparameter $\alpha$ to 0.999. For all experiments, we use the SGD optimizer, fixing the total number of training epochs to 1000 and $T_{rp}=500$, the initial learning rate at $0.001$ and continuously decay by half every 500 epochs. Each epoch is defined as cycling all labeled data once. The sinusoid spatial encoding parameter was set as $K=4$. The overall algorithm is summarized in Algorithm~\ref{alg:SSLSeg}. 

\begin{algorithm}[hb]
\SetKwData{Left}{left}\SetKwData{This}{this}\SetKwData{Up}{up}
\SetKwFunction{Concatenate}{Concatenate}\SetKwFunction{Kmeans}{K-means}
\SetKwInOut{Input}{Input}\SetKwInOut{Output}{Output}
\Input{\small{Labelled Images $\{(\matr{X}_i^l, \matr{Y}_i^l)\}$ and Unlabelled Images $\{\matr{X}_j^u\}$}}
\Output{\small{Segmentation Model Parameters $\Theta$}}
\For{$T \leftarrow 1$ \KwTo $1000$}
{
    \For{$\{\matr{X}^l_i,\matr{Y}^l_i,\matr{X}^u_j\}$ in mini-batch}
    {
        \tcp{Random Augmentation}
        Compute $t(\matr{X}^l_i), t_1(\matr{X}^u_i), t_2(\matr{X}^u_i)$\;
        \tcp{Forward Pass}
        Compute $f(t_1(\matr{X}_i^l))$,$f(t_1(\matr{X}_i^u))$,$\hat{f}(t_2(\matr{X}_i^u))$
    }
    \tcp{Supervised Loss}
    Compute $l_{dice}$ according to Eq.~(\ref{eq:DiceLoss})\;
    \tcp{Unsupervised Loss}
    Compute $l_\textrm{N-pair}$ according to Eq.~(\ref{eq:N-pairLoss})\;
    \tcp{Unlabelled Loss Weight}
    $\gamma=\exp(-10(\frac{min(T,T_{rp})}{T_{rp}}-1)^2)$\;
	\tcp{Train One Minibatch}
	$\Theta=\Theta - \alpha\nabla_\Theta l_{total}$\;
}
\caption{\small{Semi-Supervised Curvilinear Structure Segmentation: Training Algorithm}\label{alg:SSLSeg}}    

\end{algorithm}
\vspace{-0.5cm}



\section{Experiments}
\label{sec_exp}
We validate the efficacy of SemiCurv on 6 different curvilinear structure datasets spanning 3 domains: road crack segmentation, road segmentation from satellite images, and biomedical image segmentation.

\subsection{Datasets}
\noindent\textbf{CrackForest} \cite{shi2016automatic} was proposed for crack segmentation from paved road images. It consists of 118 labelled images in total. We created a datasplit for evaluating semi-supervised segmentation by randomly splitting the whole dataset into $80\%$ for training, $10\%$ for validation, and $10\%$ for testing. Among the training images, we further assume $5\%$ or $1\%$ of the images are labelled. 
\noindent\textbf{Crack500} \cite{yang2019feature} is a more comprehensive dataset consisting of 1896/348/1124 labelled road crack images for training, validation, and testing respectively. We follow the standard datasplit proposed in \cite{yang2019feature} and assume the training set is partially labelled.
\noindent\textbf{Gaps384} \cite{yang2019feature} was created from a large road crack detection dataset GAPS \cite{eisenbach2017how} by manually selecting 384 images and cropping out smaller regions with cracks, totalling 508 annotated images. We follow the standard data split for evaluation.
\noindent\textbf{MIT Road} \cite{MnihThesis} was proposed for automatically extracting road from satellite images. As we notice there are many blank regions in the RGB images, we preprocess the images to manually crop non-blank regions. In total, we obtain 6880/1215/440 labelled images for training, validation, and testing respectively.
\noindent\textbf{EM128} \cite{arganda2015crowdsourcing} was created for evaluating segmentation on cell membranes. It consists of 30 labelled images and we follow the practice in \cite{seyedhosseini2013image} to reserve 15 images for training and the rest for testing.  To allow evaluation of semi-supervised learning, we divide each image into 16 regions each covering $128\times 128$ pixels. In total we have 240/240 for training and testing respectively and we name the derived dataset as EM128.
\noindent\textbf{DRIVE128} The DRIVE dataset \cite{staal2004ridge} was developed for evaluating segmentation on retinal blood vessels. It consists of 20 labelled images which we split into 10 for training and 10 for testing. Non-overlapping patches of size $128\times 128$ pixels are cropped in a similar way to EM128 and we denote this derived dataset as DRIVE128.
The overall number of training, validation, and test samples are shown in Table~\ref{tab:SSLDataSplit}. The number of labelled training data at different levels of supervision are provided as well.
We note that the level of class imbalance is severe on most of these datasets, as can be seen from the low percentage of positive (foreground) pixels given in Table~\ref{tab:PosRatio}. The high class imbalance potentially makes the model more likely to collapse to predicting all pixels as background on the unlabelled data.

\begin{table}[htbp]
  \centering
  \caption{Percentage ($\%$) of positive pixels in datasets.}
     \setlength\tabcolsep{3pt} 
  \resizebox{0.99\linewidth}{!}{
    \begin{tabular}{ccccccc}
    \toprule
    Dataset & CrackForest & Crack500 & Gaps384 & MITRoad & EM128 & DRIVE128 \\
    \midrule
    Pos. pixel $\%$  & $2.5 \%$   & $6.0 \%$   & $1.2 \%$   & $5.1 \%$   & $22.0 \%$  & $8.6 \%$ \\
    \bottomrule
    \end{tabular}%
    }
  \label{tab:PosRatio}%
  \vspace{-0.5cm}
\end{table}%




\begin{table*}[htbp]
  \centering
  \caption{Semi-supervised learning data splits for all 6 datasets.}
  \setlength\tabcolsep{3pt}
   \resizebox{0.65\linewidth}{!}{
\begin{tabular}{cr|cc|cc|cc|cc|cc|cc}
\toprule
    Split &       & \multicolumn{2}{c|}{CrackForest} & \multicolumn{2}{c|}{Crack500} & \multicolumn{2}{c|}{Gaps384} & \multicolumn{2}{c|}{DRIVE128} & \multicolumn{2}{c|}{MITRoad} & \multicolumn{2}{c}{EM128} \\
    \midrule
    \multirow{3}[2]{*}{Train} & \multicolumn{1}{l|}{Labelled Percent.} & \multicolumn{1}{l}{5\%} & \multicolumn{1}{l|}{1\%} & \multicolumn{1}{l}{5\%} & \multicolumn{1}{l|}{1\%} & \multicolumn{1}{l}{5\%} & \multicolumn{1}{l|}{1\%} & \multicolumn{1}{l}{5\%} & \multicolumn{1}{l|}{1\%} & \multicolumn{1}{l}{1\%} & \multicolumn{1}{l|}{0.1\%} & \multicolumn{1}{l}{10\%} & \multicolumn{1}{l}{5\%} \\
          & \multicolumn{1}{l|}{\#Labeled} & \multicolumn{1}{l}{5} & \multicolumn{1}{l|}{1} & \multicolumn{1}{l}{95} & \multicolumn{1}{l|}{19} & \multicolumn{1}{l}{23} & \multicolumn{1}{l|}{5} & \multicolumn{1}{l}{8} & \multicolumn{1}{l|}{2} & \multicolumn{1}{l}{69} & \multicolumn{1}{l|}{7} & \multicolumn{1}{l}{24} & \multicolumn{1}{l}{12} \\
          & \multicolumn{1}{l|}{\#Unlabeled} & \multicolumn{1}{l}{91} & \multicolumn{1}{l|}{95} & \multicolumn{1}{l}{1801} & \multicolumn{1}{l|}{1877} & \multicolumn{1}{l}{442} & \multicolumn{1}{l|}{460} & \multicolumn{1}{l}{152} & \multicolumn{1}{l|}{158} & \multicolumn{1}{l}{6811} & \multicolumn{1}{l|}{6873} & \multicolumn{1}{l}{216} & \multicolumn{1}{l}{228} \\
    \midrule
    Test  &       & \multicolumn{2}{c|}{11} & \multicolumn{2}{c|}{1124} & \multicolumn{2}{c|}{39} & \multicolumn{2}{c|}{160} & \multicolumn{2}{c|}{440} & \multicolumn{2}{c}{240} \\
    \bottomrule
    \end{tabular}%
    }
  \label{tab:SSLDataSplit}%
\end{table*}%


\subsection{Evaluation Metric}

To evaluate the quality of segmentation predictions, we compute Intersection over Union (IoU) using a threshold of 0.5 on prediction posteriors for all test images and report the mean IoU. 
We also evaluate the F1 measure of precision and recall at threshold 0.5. This metric treats segmentation as a binary classification task. For all 6 datasets, we set the background pixel label as 0 and foreground as 1.

\subsection{Competing Methods}
We extensively compare SemiCurv with existing fully supervised methods (with different backbone networks) and state-of-the-art generic semi-supervised learning methods, as described in the following.


\noindent\textbf{Fully Supervised Methods:}
We first investigate the state-of-the-art fully supervised curvilinear structure segmentation, edge detection, and semantic segmentation methods. This provides context of what is achievable for curvilinear segmentation tasks.
\noindent\textbf{FPHBN} \cite{yang2019feature} is the state-of-the-art method for crack segmentation; we used the reported results on Crack500 in this paper for comparison.
\noindent\textbf{HED} \cite{xie2015holistically} was proposed for detecting edges in images. Due to the similar nature of edge and linear structures defined in this work, we evaluate this backbone as a baseline.
\noindent\textbf{DeepLab v3+} \cite{chen2018encoder} is the state-of-the-art  backbone network for semantic segmentation tasks. We evaluate this network here to show that generic semantic segmentation networks do not generalize well to curvilinear structure segmentation.
\noindent\textbf{UNet} \cite{ronneberger2015u} was originally proposed for medical image segmentation. We use a variant that adds residual connections in each convolution block as the backbone network in this work. Details of the backbone network is given in the Supplementary Material. 


\noindent\textbf{Semi-Supervised Methods:}
We compare against state-of-the-art semi-supervised semantic segmentation, medical image segmentation, and generic SSL methods adapted to curvilinear segmentation. 
\noindent\textbf{CutMix}~\cite{french2019semi} proposed to generate random mask to mix two images and consistency is applied to between the predictions of teacher model and student model over the masked regions.
\noindent\textbf{TCSM}~\cite{li2020transformation} adapts the mean teacher framework to learn from unlabelled data. The augmentation is limited to scaling and rotation in multiples of $90^\circ$.
\noindent\textbf{VAT}~\cite{miyato2018virtual} proposed to learn the optimal augmentation by maximizing the consistency on unlabelled data. The final objective aims to minimize the pairwise consistency as well as the entropy on unlabelled data.
\noindent\textbf{cGAN}~\cite{hung2018adversarial} \change{is another line of semi-supervised semantic segmentation approach. The conditional GAN (cGAN) based method introduced a discriminator network to differentiate ground-truth segmentation mask from predicted ones and the predictions on unlabeled data can be constrained by the discriminator.}
\noindent\textbf{Mean Teacher (Baseline)} \cite{tarvainen2017mean} proposed to use a temporal ensemble, as teacher, of a learnable student model to provide pseudo labels to train the student model. MSE consistency is used between teacher and student outputs. We use this as our baseline method.
\noindent\textbf{SemiCurv} is our proposed model incorporating differentiable geometric transformations, N-pair objective, and sinusoid positional encoding.

\subsection{Quantitative Results}


We present the comparison of both fully supervised and semi-supervised methods with $5\%$ and $1\%$ labelled data in Table~\ref{tab:MainResults}; for MITRoad and EM128 we consider different labeling budgets due to the size of dataset. In the fully supervised block, \textit{UNet} ($*95\%$) and \textit{UNet} ($*90\%$) indicate the relative performance $95\%*m$ and $90\%*m$ where $m$ is the performance achieved by the fully supervised method trained with $100\%$ of the labelled data. From the extensive comparison, we make the following observations. First, the adapted \textit{UNet} is a very strong backbone network for curvilinear structure segmentation. It outperforms the state-of-the-art semantic segmentation backbone \textit{DeepLabV3+}, edge detection network \textit{HED}, and network specifically designed for crack segmentation \textit{FPHBN} on all datasets. 
Moreover, under the semi-supervised setting, with only $5\%$ and $1\%$ labelled data \textit{SemiCurv} can match and even outperform the $95\%$ and $90\%$ relative performance of fully supervised counterparts in terms of IoU and F1, respectively. By comparing to the state-of-the-art semi-supervised methods for segmentation, we still observe very significant advantages for \textit{SemiCurv}. In particular, the improvement from baseline is more significant in the lower label regime. For example, on CrackForest IoU improved $17\%$ and $2.2\%$ from the \textit{UNet} baseline trained using $1\%$ and $5\%$ labelled data respectively. The improvement on all other datasets also exhibits similar patterns with large improvements over the UNet baseline. We also observe \textit{CutMix} to be a very competitive method, in particular on the MIT Road dataset. We speculate that this is because the MIT Road dataset resembles generic semantic segmentation problems, such that it benefits from cutmix style augmentation. Finally, \textit{TCSM} produces much worse results compared to both \textit{MT} and \textit{SemiCurv}. We speculate that this is due to weak augmentation adopted in \textit{TCSM} hampering semi-supervised performance. 

\begin{table*}[htbp]
  \centering
  \caption{\change{Quantitative evaluations of semi-supervised curvilinear segmentation. All numbers are in $\%$. (*95/90\%) indicates multiplied by 95/90\%.}}
    \begin{tabular}{cl|cc|cc|cc|cc|cc|cc}
    \toprule
          &       & \multicolumn{2}{c|}{CrackForest} & \multicolumn{2}{c|}{Crack500} & \multicolumn{2}{c|}{Gaps384} & \multicolumn{2}{c|}{DRIVE128} & \multicolumn{2}{c|}{MIT Road} & \multicolumn{2}{c}{EM128} \\
    \midrule
    \multirow{8}[5]{*}{\begin{sideways}Fully Supervised\end{sideways}} & Model & IoU   & F1    & IoU   & F1    & IoU   & F1    & IoU   & F1    & IoU   & F1    & IoU   & F1 \\
\cmidrule{2-14}          & Labelled Data & \multicolumn{2}{c|}{100\%} & \multicolumn{2}{c|}{100\%} & \multicolumn{2}{c|}{100\%} & \multicolumn{2}{c|}{100\%} & \multicolumn{2}{c|}{100\%} & \multicolumn{2}{c}{100\%} \\
\cmidrule{2-14}          & DeepLabV3 & 51.2  & 67.1  & 38.8  & 52.5  & 37.3  & 53.1  & 30.5  & 33.6  & 33.2  & 48.3  & 46.0  & 61.8 \\
          & HED   & 54.7  & 70.2  & 39.5  & 53.9  & 35.2  & 49.0  & 33.6  & 37.6  & 40.1  & 55.0  & 46.6  & 63.0 \\
          & FPHBN & NA    & NA    & 48.9  & NA    & NA    & NA    & NA    & NA    & NA    & NA    & NA    & NA \\
          & Unet 100\% & 70.7  & 84.7  & 49.6  & 62.9  & 39.0  & 52.9  & 59.2  & 60.4  & 58.4  & 71.3  & 67.5  & 80.2 \\
          & Unet 95\%  & 67.1  & 80.4  & 47.1  & 59.7  & 37.1  & 50.3  & 56.2  & \multicolumn{1}{c}{57.3} & 55.5  & \multicolumn{1}{c}{67.7} & 64.1  & 76.2 \\
          & Unet 90\% & 63.6  & 76.4  & 44.6  & 56.6  & 35.1  & 47.8  & 53.3  & 54.5  & 52.6  & 64.3  & 60.7  & 72.4 \\
          \cmidrule{2-14}
          & Labelled Data & \multicolumn{2}{c|}{5\%} & \multicolumn{2}{c|}{5\%} & \multicolumn{2}{c|}{5\%} & \multicolumn{2}{c|}{5\%} & \multicolumn{2}{c|}{1\%} & \multicolumn{2}{c}{10\%} \\
\cmidrule{2-14}    \multirow{15}[6]{*}{\begin{sideways}Semi-Supervised\end{sideways}} & Unet  & 60.1  & 74.3  & 46.5  & 60.5  & 34.9  & 48.9  & 49.1  & 56.5  & 53.8  & 67.3  & 63.4  & 77.1 \\
          & CutMix & 66.0  & 79.3  & 44.4  & 57.5  & 33.5  & 46.9  & 50.3  & 49.5  & 55.8  & 69.0  & 62.2  & 76.3 \\
          & VAT   & 53.9  & 69.1  & 48.1  & 62.5  & 29.6  & 42.3  & 47.8  & 56.3  & 48.3  & 62.6  & 57.9  & 72.6 \\
          & TCSM  & 49.9  & 66.2  & 36.8  & 50.7  & 37.8  & 52.2  & 44.8  & 51.0  & 32.5  & 47.0  & 33.0  & 48.8 \\
          & MT    & 66.9  & 79.1  & 48.0  & 61.9  & 34.3  & 48.2  & 54.2  & \multicolumn{1}{c}{60.9} & 55.2  & 68.7  & 65.3  & 78.6 \\
          & cGAN  & 67.0  & 79.2  & 48.1  & 62.0  & 33.7  & 47.3  & 55.5  & 62.4  & 53.5  & 66.6  & 62.1  & 74.8 \\
          & Ours  & \textbf{68.0} & \textbf{80.9} & \textbf{49.2} & \textbf{62.4} & \textbf{38.0} & \textbf{53.4} & \textbf{56.0} & \textbf{61.3} & \textbf{56.0} & \textbf{69.1} & \textbf{65.7} & \textbf{79.1} \\
\cmidrule{2-14}          & Labelled Data & \multicolumn{2}{c|}{1\%} & \multicolumn{2}{c|}{1\%} & \multicolumn{2}{c|}{1\%} & \multicolumn{2}{c|}{1\%} & \multicolumn{2}{c|}{0.1\%} & \multicolumn{2}{c}{5\%} \\
\cmidrule{2-14}          & Unet  & 36.6  & 48.8  & 44.4  & 58.3  & 28.3  & 41.2  & 34.1  & 45.5  & 35.9  & 50.5  & 57.0  & 71.5 \\
          & CutMix & 56.5  & 71.2  & 40.2  & 54.6  & 30.1  & 43.6  & 43.0  & 50.8  & 44.9  & 59.2  & 63.9  & 77.6 \\
          & VAT   & 29.3  & 41.5  & 43.3  & 58.4  & 18.8  & 17.3  & 31.4  & 41.9  & 25.4  & 38.6  & 57.1  & 72.2 \\
          & TCSM  & 30.5  & 43.9  & 41.9  & 56.2  & 28.6  & 41.7  & 28.8  & 34.6  & 31.6  & 45.4  & 28.6  & 43.6 \\
          & MT    & 61.9  & 75.9  & 45.6  & 59.8  & 31.1  & 44.3  & 37.7  & 47.3  & 46.4  & 61.1  & 63.2  & 76.9 \\
          & cGAN  & 54.6  & 66.9  & 45.4  & 59.5  & 30.1  & 42.8  & 42.6  & 53.4  & 43.6  & 57.4  & 60.3  & 73.4 \\
          & Ours  & \textbf{64.4} & \textbf{78.1} & \textbf{46.2} & \textbf{60.1} & \textbf{33.7} & \textbf{48.4} & \textbf{44.3} & \textbf{51.5} & \textbf{47.7} & \textbf{62.2} & \textbf{64.1} & \textbf{77.9} \\
    \bottomrule
    \end{tabular}%
  \label{tab:MainResults}%
\end{table*}%

\subsection{Ablation Study}


Here we carry out ablation studies to investigate the effectiveness of each component and present the results in Table~\ref{tab:Ablation}. 

\noindent\textbf{Supervised Loss:}
We first compare the supervised loss, \textit{SupLoss}, adopted for labelled images. \textit{UNet} with weighted binary cross entropy loss, \textit{WBCE}, is consistently worse than \textit{UNet} with dice loss, \textit{Dice}. This suggests that optimizing dice loss is able to better generalize under severe class imbalance. 

\noindent\textbf{Semi-Supervised Learning:}
We next compare the fully supervised baseline with the mean teacher model, \textit{MT}. With mean square error (MSE) consistency loss, we observe significant improvement of \textit{MT} over the baseline. We further evaluate \textit{MT} with and without the differentiable geometric transformation, \textit{GeoTform}. The significant difference in performance between the two settings indicates that strong augmentation is vital to the success of consistency-based SSL.

\begin{figure}[!htbp]
    \centering
    \subfloat[]{
    \includegraphics[width=0.49\linewidth]{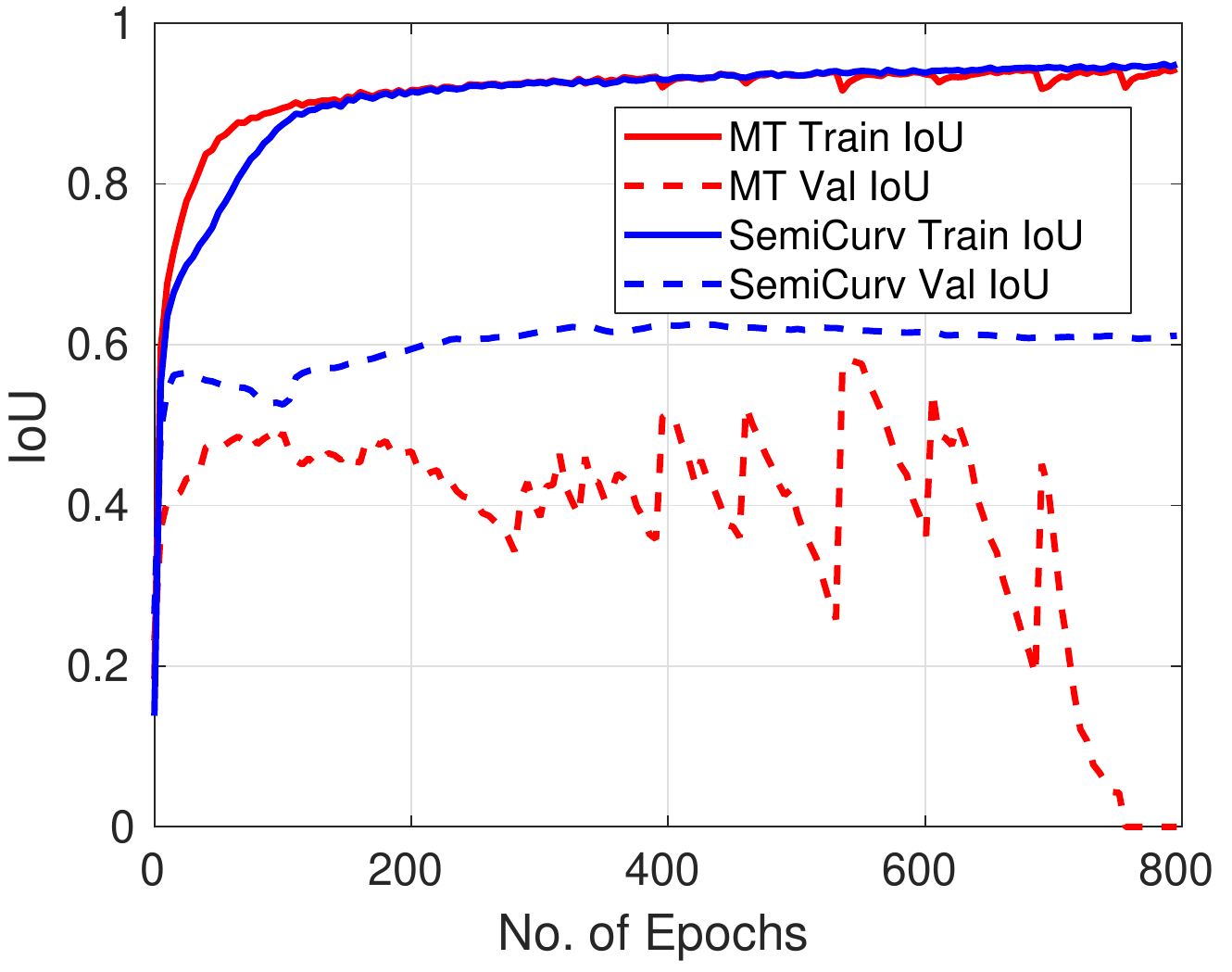}}
    \subfloat[]{\includegraphics[width=0.49\linewidth]{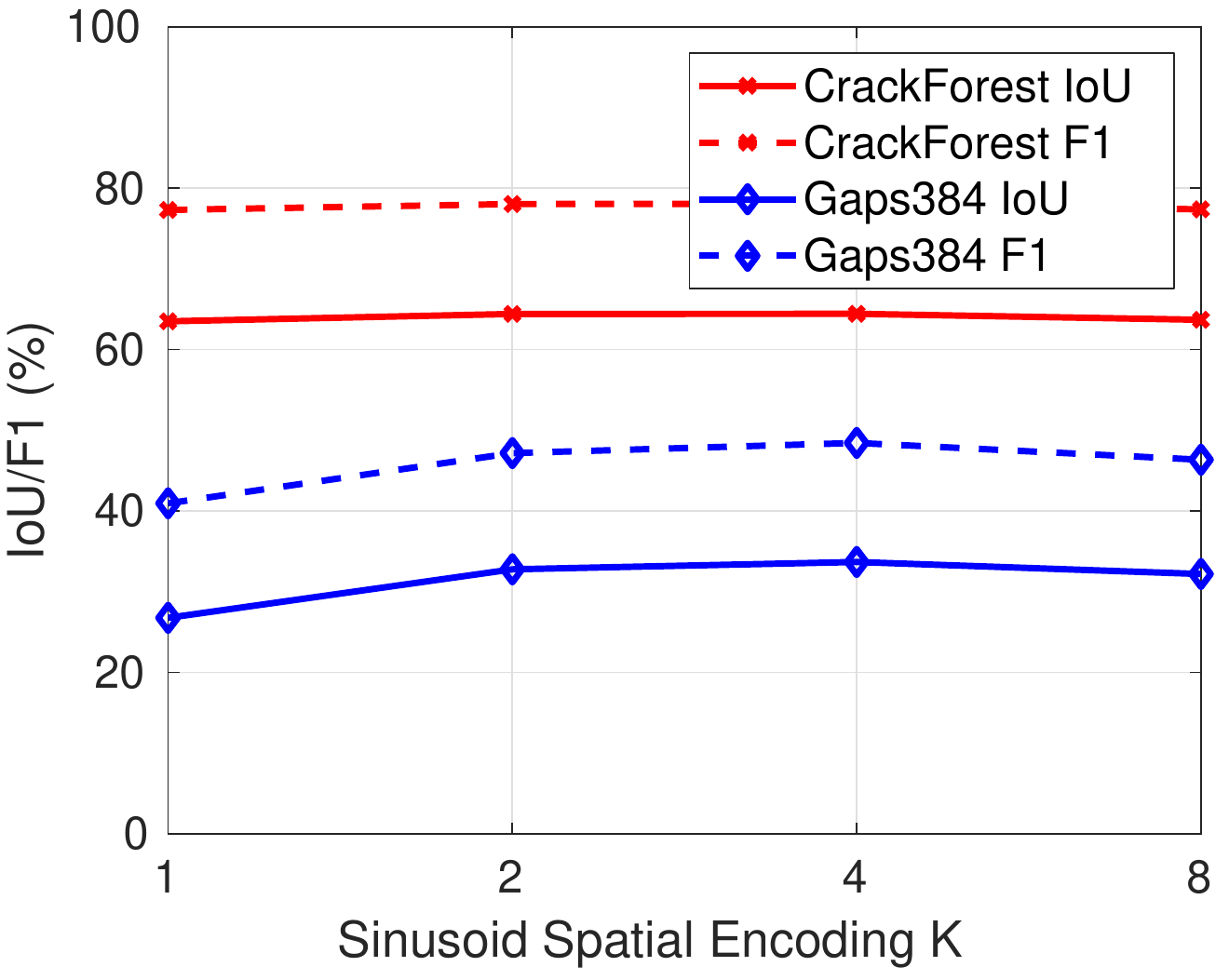}}
    \vspace{-0.3cm}
    \caption{(a)~Comparing MSE loss v.s. N-pair loss over collapsed predictions. (b)~Performance v.s. sinusoid parameter $K$.}
        \vspace{-0.3cm}
    \label{fig:CollapseSinusoidPeriod}
\end{figure}

\noindent\textbf{MSE vs N-pair Consistency Loss:}
We further compare \textit{MT} with mean square error loss, \textit{MSE}, and N-pair loss. The improvement demonstrates the superiority of considering both positive and negative pairs simultaneously with \textit{N-pair}. In addition to the numerical improvement, we also show in Fig.~\ref{fig:CollapseSinusoidPeriod}~(a) that \textit{N-pair} is more robust than \textit{MSE} and avoids collapsed predictions.
As we discussed in Sect.~\ref{sect:Consistency}, the pairwise mean square error, \textit{MSE}, consistency loss is prone to a collapsing issue. This becomes severe when the data is highly imbalanced and very few labelled samples are available for training. We present examples for both \textit{MSE} consistency and \textit{N-pair} consistency on CrackForest with only $1\%$ labelled data in Fig.~\ref{fig:CollapseSinusoidPeriod}~(a). \textit{MSE} produces unstable validation performance, and it eventually collapses to all zeros around $800$ epochs. In contrast, with the proposed \textit{N-pair} loss the validation performance keeps stable throughout the training. Overfitting, due to too few labelled data, is still observed but it avoids collapsing to all zero predictions.


\noindent\textbf{Positional Encodings:} We analyze the effect of including various forms of spatial encoding. First, when including linear spatial encoding, \textit{Linear Spt. Enc.}, we sometimes observe that performance degrades. This can be attributed to the potential overfitting to absolute locations; using sinusoid spatial encoding instead, \textit{Sinusoid Spt. Enc.}, we observe clear and consistent improvements on all datasets.
We investigate the impact of constant $K$ in Eq.~\ref{eq:SinEnc} which controls the period of sinusoid positional encoding. The final \textit{SemiCurv} is evaluated with $K=1,2,4,8$ with results shown in Fig.~\ref{fig:CollapseSinusoidPeriod}~(b). We observe an optimal range of $K$ around 4, and overall the results are robust to $K$ from $2$ to $8$. This shows sinusoid positional encoding is a robust component of our approach. 

\change{For alternative positional encodings, we evaluate two additional methods on CrackForest segmentation task. The results are presented in Tab.~\ref{tab:posembed}. We observe a small drop of performance when swapping out our proposed sinusoid encoding with learned position embedding~\cite{vaswani2017attention} and rotary position embedding~\cite{su2021roformer}. Overall, the proposed sinusoid position encoding is still the better option.}

\begin{table}[htbp]
  \centering
  \caption{Comparison of position embeddings.}
    \begin{tabular}{cccc}
    \toprule
    CrackForest & Sinusoid Enc. (ours) & Learned~\cite{vaswani2017attention} & Rotary~\cite{su2021roformer} \\
    \midrule
    1\%   & 64.41 & 62.93 & 62.96 \\
    5\%   & 68.02 & 67.49 & 67.56 \\
    \bottomrule
    \end{tabular}%
  \label{tab:posembed}%
\end{table}%

\noindent\textbf{Alternative Similarity Metric}
We further explore a different option for the similarity metric used in the N-pair loss. L2 distance is often used in N-pair loss for metric learning and we therefore evaluated the induced RBF kernel for comparison against cosine similarity. Formally, the L2 distance induced similarity is given by
\begin{equation}
    l_\textrm{N-pair}=-\frac{1}{N_B}\sum_{i=1}^{N_B}\log \frac{\exp(-||g_{i1}-\hat{g}_{i2}||_2)}{\sum_{j=1}^{N_B}\exp(-||g_{i1}-\hat{g}_{j2}||_2)}
\end{equation}

The quantitative comparison between using cosine similarity and L2 distance induced similarity is shown in Table~\ref{tab:Ablation} under the similarity metric (Sim.) column. For both similarity metrics, we use the SemiCurv framework and keep the training protocols unchanged. We observe from the comparison that cosine similarity outperforms L2 distance consistently on all datasets. Also, stability of training suffers with L2 distance compared to SemiCurv with cosine distance. This can be attributed to the normalizing effect of cosine similarity, while L2 distance is affected by the absolute number of positive pixel predictions.

\begin{table*}[htbp]
  \centering
  \caption{Ablation study for all 6 curvilinear datasets. 
  }
  \setlength\tabcolsep{1pt} 
      \resizebox{0.95\linewidth}{!}{
    \begin{tabular}{cccccccccccc}
    \toprule
          &       &       &       &       & &  CrackForest 1\% & Crack500 1\% & Gaps384 1\% & MIT Road 1\% & EM128 5\% & DRIVE128 5\% \\
    SSL   & SupLoss & GeoTform & Consist. & Spt.Enc. & Sim. & IoU   & IoU   & IoU   & IoU   & IoU   & IoU \\
    \midrule
    -     & WBCE  & \checkmark     & -     & -     & - & 21.1  & 38.6  & 24.3  & 41.7  & 40.9  & 34.9 \\
    -     & Dice  & \checkmark     & -     & -     & - & 36.6  & 44.4  & 28.3  & 51.6  & 57.0  & 49.1 \\
    MT    & Dice  & -     & MSE   & -    & - & 50.5  & 43.2  & 19.2  & 51.4  & 53.8  & 43.9 \\
    MT    & Dice  & \checkmark     & MSE   & -    & - & 61.9  & 45.6  & 31.1  & 55.2  & 63.2  & 54.2 \\
    MT    & Dice  & \checkmark     & N-pair & -   & Cosine  & 63.7  & 45.9  & 31.6  & 55.2  & 63.9  & 55.9 \\
    MT    & Dice  & \checkmark     & N-pair & Linear & Cosine & 62.8  & 45.9  & \textbf{34.0} & 54.5  & 63.7  & 44.9 \\
    MT    & Dice  & \checkmark     & N-pair & Sinusoid & Cosine & \textbf{64.4} & \textbf{46.2} & 33.7  & \textbf{56.0} & \textbf{64.1} & \textbf{56.0} \\
    MT    & Dice  & \checkmark     & N-pair & Sinusoid & L2 & 58.4 & 45.2 & 29.5  & 53.0 & 55.9 & 43.0 \\
    \bottomrule
    \end{tabular}%
    }
  \label{tab:Ablation}%
\end{table*}%

\noindent\textbf{Time Complexity}
\change{We analyze the time complexity of proposed semi-supervised learning method. With more unlabeled data, it takes more time to allow the network to iterate unlabeled data. The training time is roughly linearly proportional to the ratio of unlabeled and labeled data. We further empirically evaluated the training time for fully supervised learning method and SemiCurv in Tab.~\ref{tab:timecomplexity}. Despite more training time required, our SemiCurv performs substantially better at inference stage and there is no additional cost at inference stage.}

\begin{table}[!htbp]
  \centering
  \caption{Empirical evaluation of time complexity for proposed method.}
    \begin{tabular}{ccc}
    \toprule
    CrackForest & FullSup & SemiCurv \\
    \midrule
    1\%   & 20m   & 1h21m \\
    5\%   & 100m   & 5h30m \\
    \bottomrule
    \end{tabular}%
  \label{tab:timecomplexity}%
\end{table}%

\begin{figure}[!htb]
    \centering
    \includegraphics[width=0.8\linewidth]{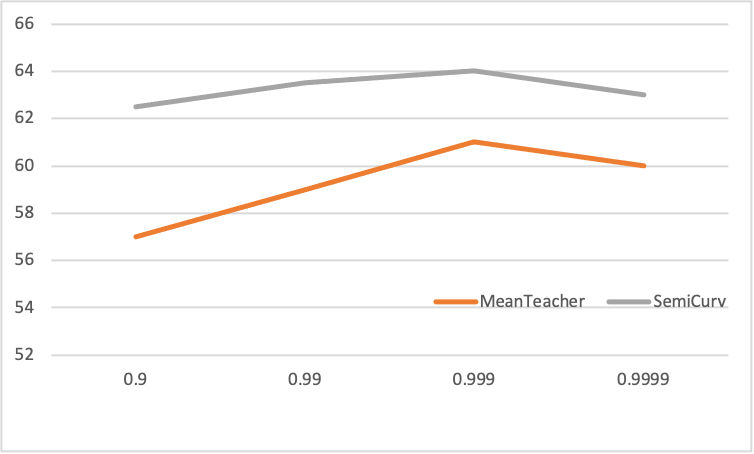}
    \caption{Evaluation of EMA hyperparameter.}
    \label{fig:hyper-alpha}
\end{figure}

\noindent\textbf{Exponential Moving Averaging Hyperparameter}
\change{We choose the EMA hyperparameter for teacher model according to the validation performance. A lower $\alpha$ will allow faster update of teacher network while a higher $\alpha$ slows down the update of teacher network. This hyperparameter is particularly sensitive for MSE consistency loss according because MSE loss is more likely to produce collapsed prediction, i.e. predicting all pixels as background. In this section we further provide empirical evidence for the chosen hyperparameter $\alpha$ by varying $\alpha$ from 0.9 to 0.9999 and compare the performance on CrackForest dataset. The results in Fig.~\ref{fig:hyper-alpha} suggest choosing $\alpha=0.999$ is the optimal choice for both mean teacher and our SemiCurv models.}

\subsection{Qualitative Results}

\begin{figure*}[!htbp]
    \centering
    \includegraphics[width=0.99\linewidth]{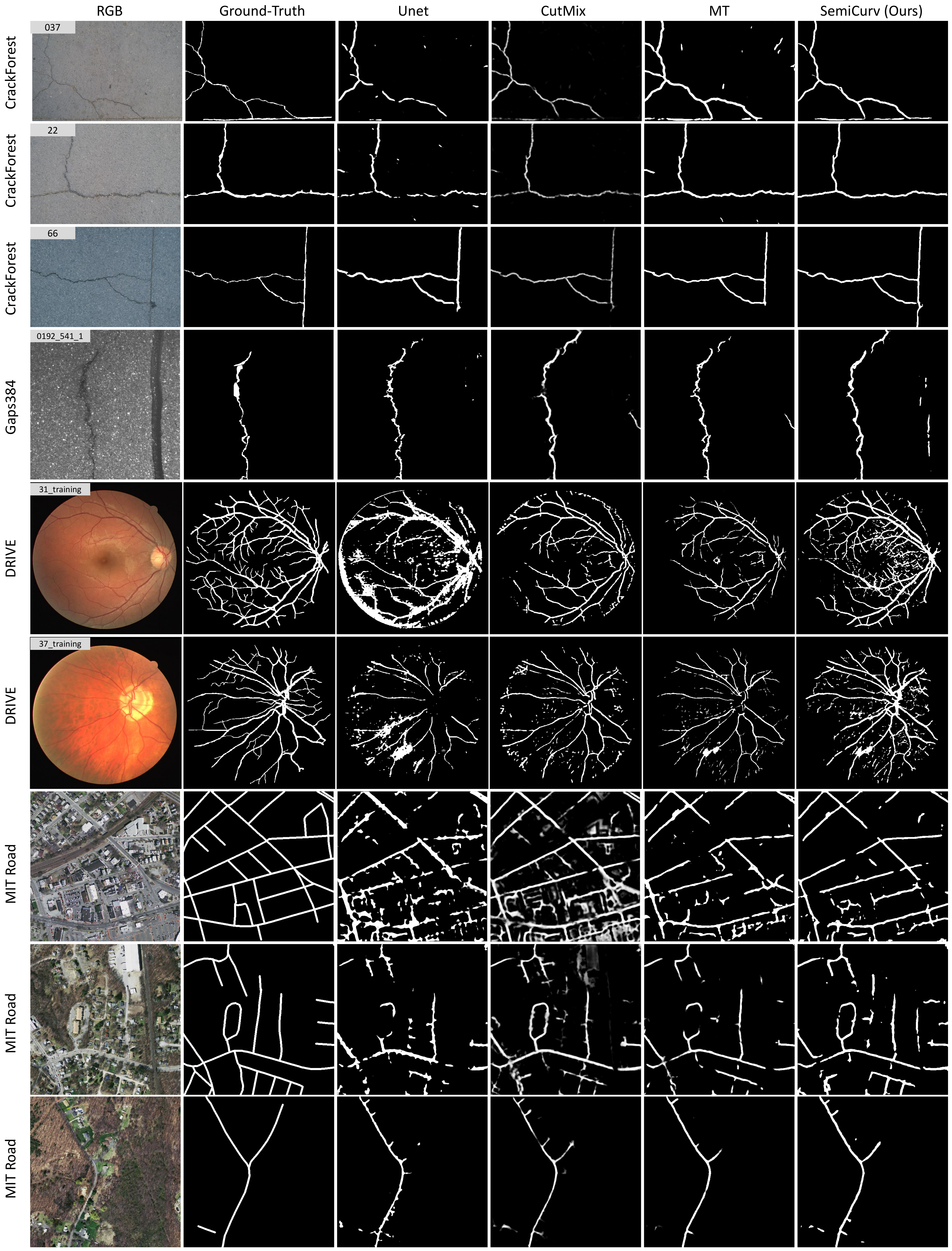}
    \caption{Qualitative comparisons for curvilinear structure segmentation.}
    \label{fig:Qualitative}
\end{figure*}

We further present qualitative comparisons of \textit{UNet}, \textit{CutMix}, \textit{MT}, and \textit{SemiCurv} on 4 datasets in Fig.~\ref{fig:Qualitative}. We make the following observations: first, for both \textit{MT} and \textit{SemiCurv}, the visual quality of segmentation results are more appealing and better mimicks the ground-truth compared to the fully supervised baseline \textit{UNet}. In particular, our method is able to generate more well-connected predictions for $037$ (1st row) 
thanks to the contribution of appending positional encoding. 
We further notice that \textit{SemiCurv} produces, in general, cleaner outputs for $037$ (1st row), $022$ (2nd row), $066\_1$ (3rd row). We also observe, in $0192\_541\_1$ (4th row), the proposed approach captures very subtle cracks on the top.
The segmentation results on DRIVE dataset (5th/6th rows) also suggest the superiority of \textit{SemiCurv}. 
It produces relatively high fidelity and well-delineated blood vessel segmentation maps. In comparison, \textit{UNet}, \textit{CutMix}, and \textit{MT} tend to mingle different blood vessels together which is less common in \textit{SemiCurv}'s predictions.
For satellite image segmentation, the visualization covers industrial areas (row 7) and rural areas (rows 8-9). We observe consistent improvements of \textit{SemiCurv} compared to both \textit{UNet} and state-of-the-art semi-supervised image segmentation \textit{CutMix}. Interestingly, the \textit{SemiCurv} predictions sometimes contain valid road branches that are missing in the ground-truth annotation (last row).


\section{Conclusion}
In this work we addressed curvilinear structure segmentation in a semi-supervised learning setting to exploit ``freely'' available and abundant amounts of unlabelled data with our proposed SemiCurv framework. In particular, we introduced stronger augmentation involving affine geometric transformations which we showed to be essential to the success of SSL. We further identified implications of the severe class imbalance in curvilinear segmentation tasks on the widely used MSE consistency loss, and showed that N-pair loss should be used instead to mitigate these issues. Finally, we found that sinusoid positional encoding is effective in further improving segmentation performance. Our extensive experiments on 6 curvilinear structure segmentation datasets from 3 different domains demonstrates the effectiveness of SemiCurv, and detailed ablation studies validated the importance of each of our proposed components. Our SemiCurv framework and the extensive results presented in this work provide a foundation for future work in SSL for curvilinear segmentation.


%

\appendices

\section*{Acknowledgment}
This research is supported by the Agency for Science, Technology and Research (A*STAR) under its AME Programmatic Funds (Grant No. A20H6b0151).


\ifCLASSOPTIONcaptionsoff
  \newpage
\fi



%
\bibliographystyle{IEEEtran}
\bibliography{reference}

\begin{thebibliography}{10}
\providecommand{\url}[1]{#1}
\csname url@samestyle\endcsname
\providecommand{\newblock}{\relax}
\providecommand{\bibinfo}[2]{#2}
\providecommand{\BIBentrySTDinterwordspacing}{\spaceskip=0pt\relax}
\providecommand{\BIBentryALTinterwordstretchfactor}{4}
\providecommand{\BIBentryALTinterwordspacing}{\spaceskip=\fontdimen2\font plus
\BIBentryALTinterwordstretchfactor\fontdimen3\font minus
  \fontdimen4\font\relax}
\providecommand{\BIBforeignlanguage}[2]{{%
\expandafter\ifx\csname l@#1\endcsname\relax
\typeout{** WARNING: IEEEtran.bst: No hyphenation pattern has been}%
\typeout{** loaded for the language `#1'. Using the pattern for}%
\typeout{** the default language instead.}%
\else
\language=\csname l@#1\endcsname
\fi
#2}}
\providecommand{\BIBdecl}{\relax}
\BIBdecl

\bibitem{yang2019feature}
F.~Yang, L.~Zhang, S.~Yu, D.~Prokhorov, X.~Mei, and H.~Ling, ``Feature pyramid
  and hierarchical boosting network for pavement crack detection,'' \emph{IEEE
  Transactions on Intelligent Transportation Systems}, 2019.

\bibitem{wang2019context}
F.~Wang, Y.~Gu, W.~Liu, Y.~Yu, S.~He, and J.~Pan, ``Context-aware
  spatio-recurrent curvilinear structure segmentation,'' in \emph{IEEE
  Conference on Computer Vision and Pattern Recognition}, 2019.

\bibitem{hu2019topology}
X.~Hu, F.~Li, D.~Samaras, and C.~Chen, ``Topology-preserving deep image
  segmentation,'' in \emph{Advances in Neural Information Processing Systems},
  2019.

\bibitem{mosinska2018beyond}
A.~Mosinska, P.~Marquez-Neila, M.~Kozi{\'n}ski, and P.~Fua, ``Beyond the
  pixel-wise loss for topology-aware delineation,'' in \emph{IEEE Conference on
  Computer Vision and Pattern Recognition}, 2018.

\bibitem{van2020survey}
J.~E. Van~Engelen and H.~H. Hoos, ``A survey on semi-supervised learning,''
  \emph{Machine Learning}, vol. 109, no.~2, pp. 373--440, 2020.

\bibitem{hung2019adversarial}
W.~C. Hung, Y.~H. Tsai, Y.~T. Liou, Y.~Y. Lin, and M.~H. Yang, ``Adversarial
  learning for semi-supervised semantic segmentation,'' in \emph{British
  Machine Vision Conference}, 2018.

\bibitem{mondal2019revisiting}
A.~K. Mondal, A.~Agarwal, J.~Dolz, and C.~Desrosiers, ``Revisiting cyclegan for
  semi-supervised segmentation,'' \emph{arXiv preprint arXiv:1908.11569}, 2019.

\bibitem{ouali2020semi}
Y.~Ouali, C.~Hudelot, and M.~Tami, ``Semi-supervised semantic segmentation with
  cross-consistency training,'' in \emph{IEEE/CVF Conference on Computer Vision
  and Pattern Recognition}, 2020.

\bibitem{french2019semi}
G.~French, T.~Aila, S.~Laine, M.~Mackiewicz, and G.~Finlayson,
  ``Semi-supervised semantic segmentation needs strong, high-dimensional
  perturbations,'' in \emph{British Machine Vision Conference}, 2020.

\bibitem{olsson2021classmix}
V.~Olsson, W.~Tranheden, J.~Pinto, and L.~Svensson, ``Classmix:
  Segmentation-based data augmentation for semi-supervised learning,'' in
  \emph{Proceedings of the IEEE/CVF Winter Conference on Applications of
  Computer Vision}, 2021.

\bibitem{everingham2015pascal}
M.~Everingham, S.~A. Eslami, L.~Van~Gool, C.~K. Williams, J.~Winn, and
  A.~Zisserman, ``The pascal visual object classes challenge: A
  retrospective,'' \emph{International journal of computer vision}, 2015.

\bibitem{andrew2001multiple}
A.~M. Andrew, ``Multiple view geometry in computer vision,'' \emph{Kybernetes},
  2001.

\bibitem{sohn2016improved}
K.~Sohn, ``Improved deep metric learning with multi-class n-pair loss
  objective,'' in \emph{Advances in neural information processing systems},
  2016.

\bibitem{chen2020simple}
T.~Chen, S.~Kornblith, M.~Norouzi, and G.~Hinton, ``A simple framework for
  contrastive learning of visual representations,'' in \emph{International
  Conference on Machine Learning}, 2020.

\bibitem{he2020momentum}
K.~He, H.~Fan, Y.~Wu, S.~Xie, and R.~Girshick, ``Momentum contrast for
  unsupervised visual representation learning,'' in \emph{Proceedings of the
  IEEE/CVF Conference on Computer Vision and Pattern Recognition}, 2020.

\bibitem{liu2018intriguing}
R.~Liu, J.~Lehman, P.~Molino, F.~P. Such, E.~Frank, A.~Sergeev, and
  J.~Yosinski, ``An intriguing failing of convolutional neural networks and the
  coordconv solution,'' in \emph{Advances in Neural Information Processing
  Systems}, 2018.

\bibitem{mnih2010learning}
V.~Mnih and G.~E. Hinton, ``Learning to detect roads in high-resolution aerial
  images,'' in \emph{European Conference on Computer Vision}, 2010.

\bibitem{MnihThesis}
V.~Mnih, ``Machine learning for aerial image labeling,'' Ph.D. dissertation,
  University of Toronto, 2013.

\bibitem{henry2018road}
C.~Henry, S.~M. Azimi, and N.~Merkle, ``Road segmentation in sar satellite
  images with deep fully convolutional neural networks,'' \emph{IEEE Geoscience
  and Remote Sensing Letters}, 2018.

\bibitem{liu2018roadnet}
Y.~Liu, J.~Yao, X.~Lu, M.~Xia, X.~Wang, and Y.~Liu, ``Roadnet: Learning to
  comprehensively analyze road networks in complex urban scenes from
  high-resolution remotely sensed images,'' \emph{IEEE Transactions on
  Geoscience and Remote Sensing}, 2018.

\bibitem{ronneberger2015u}
O.~Ronneberger, P.~Fischer, and T.~Brox, ``U-net: Convolutional networks for
  biomedical image segmentation,'' in \emph{International Conference on Medical
  image computing and computer-assisted intervention}.\hskip 1em plus 0.5em
  minus 0.4em\relax Springer, 2015.

\bibitem{isensee2019nnu}
F.~Isensee, J.~Petersen, A.~Klein, D.~Zimmerer, P.~F. Jaeger, S.~Kohl,
  J.~Wasserthal, G.~Koehler, T.~Norajitra, S.~Wirkert \emph{et~al.}, ``nnu-net:
  Self-adapting framework for u-net-based medical image segmentation,'' in
  \emph{Bildverarbeitung f{\"u}r die Medizin 2019}.\hskip 1em plus 0.5em minus
  0.4em\relax Springer, 2019.

\bibitem{arganda2015crowdsourcing}
I.~Arganda-Carreras, S.~C. Turaga, D.~R. Berger, D.~Cire{\c{s}}an, A.~Giusti,
  L.~M. Gambardella, J.~Schmidhuber, D.~Laptev, S.~Dwivedi, J.~M. Buhmann
  \emph{et~al.}, ``Crowdsourcing the creation of image segmentation algorithms
  for connectomics,'' \emph{Frontiers in neuroanatomy}, 2015.

\bibitem{zou2018deepcrack}
Q.~Zou, Z.~Zhang, Q.~Li, X.~Qi, Q.~Wang, and S.~Wang, ``Deepcrack: Learning
  hierarchical convolutional features for crack detection,'' \emph{IEEE
  Transactions on Image Processing}, 2018.

\bibitem{laine2016temporal}
S.~Laine and T.~Aila, ``Temporal ensembling for semi-supervised learning,''
  \emph{International Conference on Learning Representations}, 2017.

\bibitem{tarvainen2017mean}
A.~Tarvainen and H.~Valpola, ``Mean teachers are better role models:
  Weight-averaged consistency targets improve semi-supervised deep learning
  results,'' in \emph{Advances in neural information processing systems}, 2017.

\bibitem{miyato2018virtual}
T.~Miyato, S.-i. Maeda, M.~Koyama, and S.~Ishii, ``Virtual adversarial
  training: a regularization method for supervised and semi-supervised
  learning,'' \emph{IEEE transactions on pattern analysis and machine
  intelligence}, 2018.

\bibitem{rasmus2015semi}
A.~Rasmus, M.~Berglund, M.~Honkala, H.~Valpola, and T.~Raiko, ``Semi-supervised
  learning with ladder networks,'' in \emph{Advances in neural information
  processing systems}, 2015.

\bibitem{goodfellow2014explaining}
I.~J. Goodfellow, J.~Shlens, and C.~Szegedy, ``Explaining and harnessing
  adversarial examples,'' \emph{arXiv preprint arXiv:1412.6572}, 2014.

\bibitem{berthelot2019mixmatch}
D.~Berthelot, N.~Carlini, I.~Goodfellow, N.~Papernot, A.~Oliver, and C.~A.
  Raffel, ``Mixmatch: A holistic approach to semi-supervised learning,'' in
  \emph{Advances in Neural Information Processing Systems}, 2019.

\bibitem{odena2016semi}
A.~Odena, ``Semi-supervised learning with generative adversarial networks,''
  \emph{arXiv preprint arXiv:1606.01583}, 2016.

\bibitem{souly2017semi}
N.~Souly, C.~Spampinato, and M.~Shah, ``Semi supervised semantic segmentation
  using generative adversarial network,'' in \emph{IEEE International
  Conference on Computer Vision}, 2017.

\bibitem{li2020transformation}
X.~Li, L.~Yu, H.~Chen, C.-W. Fu, L.~Xing, and P.-A. Heng,
  ``Transformation-consistent self-ensembling model for semisupervised medical
  image segmentation,'' \emph{IEEE Transactions on Neural Networks and Learning
  Systems}, 2020.

\bibitem{xie2020unsupervised}
Q.~Xie, Z.~Dai, E.~Hovy, T.~Luong, and Q.~Le, ``Unsupervised data augmentation
  for consistency training,'' in \emph{Advances in Neural Information
  Processing Systems}, 2020, pp. 6256--6268.

\bibitem{vaswani2017attention}
A.~Vaswani, N.~Shazeer, N.~Parmar, J.~Uszkoreit, L.~Jones, A.~N. Gomez,
  {\L}.~Kaiser, and I.~Polosukhin, ``Attention is all you need,'' in
  \emph{Advances in Neural Information Processing Systems}, 2017.

\bibitem{4667275}
X.~{Guo}, Y.~{Yin}, C.~{Dong}, G.~{Yang}, and G.~{Zhou}, ``On the class
  imbalance problem,'' in \emph{2008 Fourth International Conference on Natural
  Computation}, 2008.

\bibitem{johnson2019survey}
J.~M. Johnson and T.~M. Khoshgoftaar, ``Survey on deep learning with class
  imbalance,'' \emph{Journal of Big Data}, 2019.

\bibitem{Guo2017LearningFC}
H.~Guo, Y.~Li, J.~Shang, G.~Mingyun, H.~Yuanyue, and G.~Bing, ``Learning from
  class-imbalanced data: Review of methods and applications,'' \emph{Expert
  Syst. Appl.}, pp. 220--239, 2017.

\bibitem{sudre2017generalised}
C.~H. Sudre, W.~Li, T.~Vercauteren, S.~Ourselin, and M.~J. Cardoso,
  ``Generalised dice overlap as a deep learning loss function for highly
  unbalanced segmentations,'' in \emph{Deep learning in medical image analysis
  and multimodal learning for clinical decision support}.\hskip 1em plus 0.5em
  minus 0.4em\relax Springer, 2017.

\bibitem{shi2016automatic}
Y.~Shi, L.~Cui, Z.~Qi, F.~Meng, and Z.~Chen, ``Automatic road crack detection
  using random structured forests,'' \emph{IEEE Transactions on Intelligent
  Transportation Systems}, 2016.

\bibitem{eisenbach2017how}
M.~Eisenbach, R.~Stricker, D.~Seichter, K.~Amende, K.~Debes, M.~Sesselmann,
  D.~Ebersbach, U.~Stoeckert, and H.-M. Gross, ``How to get pavement distress
  detection ready for deep learning? a systematic approach.'' in
  \emph{International Joint Conference on Neural Networks}, 2017.

\bibitem{seyedhosseini2013image}
M.~Seyedhosseini, M.~Sajjadi, and T.~Tasdizen, ``Image segmentation with
  cascaded hierarchical models and logistic disjunctive normal networks,'' in
  \emph{IEEE International Conference on Computer Vision}, 2013.

\bibitem{staal2004ridge}
J.~Staal, M.~D. Abr{\`a}moff, M.~Niemeijer, M.~A. Viergever, and
  B.~Van~Ginneken, ``Ridge-based vessel segmentation in color images of the
  retina,'' \emph{IEEE transactions on medical imaging}, 2004.

\bibitem{xie2015holistically}
S.~Xie and Z.~Tu, ``Holistically-nested edge detection,'' in \emph{IEEE
  international conference on computer vision}, 2015.

\bibitem{chen2018encoder}
L.-C. Chen, Y.~Zhu, G.~Papandreou, F.~Schroff, and H.~Adam, ``Encoder-decoder
  with atrous separable convolution for semantic image segmentation,'' in
  \emph{European Conference on Computer Vision}, 2018.

\bibitem{hung2018adversarial}
W.-C. Hung, Y.-H. Tsai, Y.-T. Liou, Y.-Y. Lin, and M.-H. Yang, ``Adversarial
  learning for semi-supervised semantic segmentation,'' in \emph{British
  Machine Vision Conference}, 2018.

\bibitem{su2021roformer}
J.~Su, Y.~Lu, S.~Pan, B.~Wen, and Y.~Liu, ``Roformer: Enhanced transformer with
  rotary position embedding,'' \emph{arXiv preprint arXiv:2104.09864}, 2021.

\end{thebibliography}

%








\end{document}